\documentclass{article}

    \usepackage[final]{neurips_data_2021}

\usepackage[utf8]{inputenc} 
\usepackage[T1]{fontenc}    
\usepackage{hyperref}       
\usepackage{url}            
\usepackage{booktabs}       
\usepackage{amsfonts}       
\usepackage{nicefrac}       
\usepackage{microtype}      
\usepackage{xcolor}         

\usepackage{tikz}

\newcommand{\udensdash}[1]{
    \tikz[baseline=(todotted.base)]{
        \node[inner sep=1pt,outer sep=0pt] (todotted) {#1};
        \draw[densely dashed] (todotted.south west) -- (todotted.south east);
    }
}

\newcommand{\udensdot}[1]{
    \tikz[baseline=(todotted.base)]{
        \node[inner sep=1pt,outer sep=0pt] (todotted) {#1};
        \draw[densely dotted] (todotted.south west) -- (todotted.south east);
    }
}

\usepackage{url}
\usepackage{multirow}
\usepackage{adjustbox}
\usepackage{footnote}
\usepackage{booktabs}
\usepackage{graphicx}
\usepackage[none]{hyphenat}
\usepackage{makecell}
\usepackage{framed}
\usepackage{tabto}
\usepackage{tipa}
\usepackage{subcaption}

\usepackage{amsmath}

\newcommand\unessential[1]{\textcolor{gray}{#1}}

\usepackage{ulem}
\usepackage{makecell}

\title{NATURE: \underline{N}atural \underline{A}uxiliary \underline{T}ext \underline{U}tterances for \underline{R}ealistic Spoken Language \underline{E}valuation}  

\author{
  David Alfonso-Hermelo\textsuperscript{*} \\
  \texttt{david.alfonso.hermelo@huawei.com} \\
  \And
  Ahmad Rashid\textsuperscript{*†‡} \\
  \texttt{ahmad.rashid@huawei.com} \\
  \And
  Abbas Ghaddar\textsuperscript{*} \\
  \texttt{abbas.ghaddar@huawei.com} \\
  \And
  Philippe Langlais\textsuperscript{§} \\
  \texttt{felipe@iro.umontreal.ca} \\
  \And
  Mehdi Rezagholizadeh\textsuperscript{*} \\
  \texttt{mehdi.rezagholizadeh@huawei.com} \\
  \\
  \textsuperscript{*}\small{Huawei Noah's Ark Lab} \hspace{0.5em} \textsuperscript{†}\small{Vector Institute} \hspace{0.5em} \textsuperscript{‡}\small{University of Waterloo} \hspace{0.5em} \textsuperscript{§}\small{RALI/DIRO, Université de Montréal} \\
}

\begin{document}

\maketitle

\begin{abstract}
  Slot-filling and intent detection are the backbone of conversational agents such as voice assistants, and are active areas of research. Even though state-of-the-art techniques on publicly available benchmarks show impressive performance, their ability to generalize to realistic scenarios is yet to be demonstrated. In this work, we present NATURE, a set of simple spoken-language oriented transformations, applied to the evaluation set of datasets, to introduce human spoken language variations while preserving the semantics of an utterance. We apply NATURE to common slot-filling and intent detection benchmarks and demonstrate that simple perturbations from the standard evaluation set by NATURE can deteriorate model performance significantly. Through our experiments we demonstrate that when NATURE operators are applied to evaluation set of popular benchmarks the model accuracy can drop by up to 40\%.  
\end{abstract}

\section{Introduction}
\label{sec:intro}

The past decade has seen a proliferation of voice assistants (VAs) and conversational agents in our daily lives. This has been possible due to the progress in the fields of natural language understanding (NLU), spoken language understanding (SLU) and natural language processing (NLP). Commercial VAs are typically pipeline systems with an NLU engine which attempts to categorize and understand user intent. The main component of the NLU engine is a slot-filling (SF) and intent detection (ID) model. State-of-the-art models~\cite{qin2019stack,wang2018bi,yamada2020luke} generally report a high accuracy and F1 score on popular benchmarks such as ATIS~\cite{hemphill1990atis} or SNIPS~\cite{coucke2018SNIPS} which may give an impression that the problem is solved. However, these benchmarks do not model the distinctive variations of spoken-language and the characteristics that a VA must handle in real scenarios. 

It has been observed across the fields of NLP and NLU that state-of-the-art deep learning models fit on the spurious, surface-level patterns of the datasets~\citep{mccoy2019right,zhang2019paws,jin2019bert}. A growing body of work has demonstrated this on challenging evaluation sets designed by perturbing an existing evaluation set. These perturbations include character-level additions, deletions and swaps for machine translation~\citep{belinkov2017synthetic}, character-level adversarial perturbations to trick neural classifiers~\citep{ebrahimi2017hotflip}, using a masked language model to generate adversarial perturbations~\cite{gargbae} and a heuristic based word substitution model to generate semantically plausible adversarial text~\cite{jin2020bert} among others.

Making a large, diverse, spoken-language oriented, multilingual benchmark would be ideal. However this is a labor-intensive, time-consuming, and expensive commitment. As a trade-off between less costly annotation and more realistic data, we propose a framework that focuses on perturbing the existing evaluation set by applying simple, spoken-language oriented, realistic operators\footnote{By realistic, we mean that modified utterances remain semantically similar to the original intention in a real-life scenario.} that modify the input sentence without perturbing the original meaning. In this paper, we introduce the NATURE (\textit{Natural Alterations of Textual Utterances for Realistic Evaluation}) framework, a compilation of operators that preserve the original semantics while adding realistic spoken-language characteristics to the evaluation set. In addition to producing realistic data, the score analysis of the evaluation sets perturbed with a single operator helps pinpoint the superficial and heuristic dependencies of each model. To the best of our knowledge, no work has attempted to demonstrate that the benchmarks and models for the dual tasks of SF and ID rely on frequent heuristic patterns. Table \ref{tab:bad_pred} shows examples of perturbed utterances where a state-of-the-art model \cite{qin2019stack} correctly predicted the label for the original utterance but failed for the perturbed utterance.

\begin{savenotes}
  \setlength{\tabcolsep}{3pt}
    \footnotesize
    \centering
    \begin{table}
        \resizebox{\textwidth}{!}{

        \begin{tabular}{l|ll}
        \toprule
        \bf Utterance & \bf Task: & \bf Model Prediction Errors \\
        \midrule
        
        \makecell[l]{play party anthems \\ $\rightarrow$ \textbf{ploy} party anthems} & ID: & \makecell[l]{\texttt{Play\_Music} \\ $\rightarrow$ \texttt{Search\_Creative\_Work}} \\
        \\
        
        \makecell[l]{play some sixties music \\ $\rightarrow$ \textbf{plays} some sixties music} & SF: & \makecell[l]{[sixties]:\texttt{year} \\ $\rightarrow$ [sixties]:\texttt{year}; [plays]:\texttt{album}}\\
        \\
        
        \multirow{2}{*}{\makecell[l]{listen to dragon ball: music collection \\ $\rightarrow$ \textbf{like} listen to dragon ball: music collection}} & ID: & \makecell[l]{\texttt{Search\_Creative\_Work} \\ $\rightarrow$ \texttt{Play\_Music}\vspace{2mm}}\\
        & SF: & \makecell[l]{[dragon ball: music collection]:\texttt{object\_name} \\ $\rightarrow$ [dragon ball]:\texttt{artist}; [collection]:\texttt{album}}\\
        \bottomrule
        \end{tabular}
        }
    \caption{Examples of NATURE-perturbed utterances with badly predicted slots and/or intent. The perturbed utterance is preceded by a $\rightarrow$. }
    \label{tab:bad_pred}
    \end{table}
\end{savenotes}

\section{Related Work}
\label{sec:related_work}

\subsection{Realizing models use shortcuts}
A growing number of studies identify a tendency in NLU models to leverage the superficial features and language artifacts instead of generalizing over the semantic content. A naive way to force generalization is to automatically add noise to the training set, however, as demonstrated by \cite{belinkov2017synthetic}, models trained on synthetic noise do not necessarily perform well on natural noise, requiring a more elaborate approach.
Given our incapacity to control the features these models learn, each task requires an in-depth analysis and a data or model modification that guides it to the correct answer. For the political claims detection task \cite{pado2019sides} and \cite{dayanik2020masking} unveil a strong bias towards the claims made by frequent actors that require masking the actor and its pronouns during training to improve the performance. 
Other works have focused on the artifact and heuristic over-fitting for the natural language inference (NLI) task~\citep{gururangan2018annotation, poliak2018hypothesis, zellers2018swag, mccoy2019right, naik2018stress} or for the question-answering (QA) task \cite{jia2017adversarial}.
The work of \cite{ghaddar2021context} focuses on the artifacts in named entity recognition task and \cite{balasubramanian2020s} shows that substituting Named Entities (NEs) influences the robustness of BERT-based models for different tasks (NLI, co-reference resolution and grammar error correction).

\subsection{Alternative evaluation}
Some researchers have proposed evaluation sets with naturally occurring adverse sentences for different tasks such as HANS for natural language inference (MNLI)~\cite{mccoy2019right} or PAWS~\cite{zhang2019paws} and PAWS-X \cite{yang2019paws} for paraphrase identification.
Another strategy involves a systematic perturbation of the evaluation set \cite{lin2020rigorous}. This has gained popularity in recent years with a growing interest in more challenging and adversarial evaluation frameworks. However, a more challenging evaluation set has to ensure high quality annotation, which is why many papers have suggested a human-in-the-loop approach \cite{kaushik2019learning, gardner2020evaluating, kiela2021dynabench}. But these approaches are costly, specially due to the number and quality of annotators necessary to produce a high-quality output.
Generalization is more easily achieved when the training data is large and diverse. A model can be effective, yet, if it is only fed with small and/or similar data, it will have difficulties to achieve robustness. Some researchers \cite{louvan2020simple, zeng2020counterfactual, dai2020analysis, min2020syntactic, moosavi2020improving} use data augmentation strategies to improve the training data and help boost a model's performance.

Other researchers follow a different path and suggest evaluating by perturbing the evaluation set using multiple task-agnostic rule-based transformations. These slightly alter the form of the data while affecting very little the semantic content. In this category, we can cite the works of \cite{ribeiro2020beyond} (Checklist tool) and \cite{goel2021robustness} (RobustnessGym).

\subsection{Spoken-Language perturbation methods}
There have been a few works that have done research on spoken-language oriented perturbation methods. Some seek to simulate automatic speech recognition (ASR) errors \cite{tsvetkov2014augmenting, simonnet2018simulating, li2018improving, gopalakrishnan2020neural}. Whether using mappings of common ASR errors or based on the acoustic word embedding approach~\cite{bengio2014word}, these strategies cannot work for SF and ID because we may loose the token-by-token semantic labeling that is required for SF.

Other works have devised methods that change the sentence form while keeping track of the semantic labeling \cite{yin2020robustness, li2020coco}; although they are not presented as spoken-language oriented. Such approaches, whether they emulate non-native speaker errors or produce counterfactual versions of the original utterances, use value-substitution techniques that require high-quality label-token dictionaries for each new dataset.

In NATURE, we aim to produce spoken-language oriented perturbations~\cite{bengio2014word, tsvetkov2014augmenting, simonnet2018simulating, li2018improving, gopalakrishnan2020neural}, such that the utterances remain semantically similar~\cite{ribeiro2020beyond, goel2021robustness}. without using costly label-token dictionaries~\cite{mccrae2020english, lin2012entity, lin2020rigorous} and human-in-the-loop techniques~\cite{kaushik2019learning, gardner2020evaluating, kiela2021dynabench}.

\section{Methodology}
\label{sec:methodo}

We divide the NATURE operators into three categories - fillers, synonyms and \textit{speako} (or similar sounding). Since these operators are intended to introduce human speech inspired small perturbations in SF and ID evaluation, it is desirable for a trained model to maintain its performance under NATURE perturbations. Table~\ref{tab:examples} gives a few examples of these operators.

\subsection{Fillers}
\label{ssec:fill}
Fillers are ubiquitous in everyday spoken language and often appear in transcribed human-to-human dialog corpora (such as the Switchboard corpus \cite{godfrey1992switchb}, composed of approximately 1.6\% fillers \cite{shriberg2001errrr}). 

Fillers serve as hesitation markers (e.g.: \textit{Bring me the\textbf{, like,} Greek yogurt. I've heard it's really\textbf{, you know,} savoury.}) or as introduction/closure of a turn of speech (e.g., \textit{\textbf{Now,} bring me the Greek yogurt \textbf{please and thank you}. \textbf{Actually,} I've heard it's really savoury\textbf{, right?}}).

Because they are semantically poor (lacking essential meaning) and therefore do not change the overall meaning of an utterance, fillers are intentionally cleaned off in SF and ID benchmarks. Although we could design a pre-processing step to remove fillers from a VA system it is more interesting to study the impact of fillers and to test the capacity of models, specially those pre-trained on language modeling, to generalize over utterances with fillers.

We propose 4 different filler operators:
\begin{itemize}
    \item \textbf{Beginning-of-sentence} (BOS): a small introductory filler phrase at the beginning of the utterance, such as: \textit{so}, \textit{like}, \textit{actually}, \textit{okay so}, \textit{so okay}, \textit{so basically}, \textit{now} or \textit{well}.
    \item \textbf{End-of-sentence} (EOS): a small conclusive filler phrase at the end of the utterance, such as: \textit{if you please},  \textit{please and thank you}, \textit{if you can}, \textit{right now}, \textit{right away}, \textit{would you mind ?}
    \item \textbf{Pre-verb}: a filler word or sequence of words appearing before the utterance's verb or verbal phrase, such as: \textit{like}, \textit{basically} or \textit{actually}.
    \item \textbf{Post-verb}: a filler word or sequence of words appearing after the utterance's verb or verbal phrase, such as: \textit{basically}, \textit{actually}, \textit{like} or \textit{you know}.
\end{itemize}

BOS and EOS operators simply add a filler at the very beginning or the end of the utterance, respectively. The pre-verb and post-verb operators require us to find the part-of-speech (POS) tag of the utterance tokens \footnote{We use the NLTK library to find the POS of the tokens.}. Then, the filler is put at the correct place. We add a fail-safe rule to ensure that a filler is added if no verb is found where expected. To that end, we use the overly-recurrent filler, \textit{like}, and the first appearing NE as a pivot instead of the first appearing verb e.g., \textit{let's check \textbf{like} avengers)}.

\subsection{Synonyms}
\label{ssec:synonym}
A synonym is a word that can be interchanged with another word in the context, without changing the meaning of the whole. To replicate this semantic operation, we select the POS corresponding to the NATURE operator (verb, adjective, adverb, etc.). Then, we select a word of that type in the input utterance and make a list of corresponding potential synonym candidates (with the same POS tag) to replace it. 
Next, we use a pre-trained BERT-base model with a language modeling head to produce corresponding probabilities of synonym candidates. We use this BERT-based model instead of a human populated dictionary (such as Wiktionary) since not all dictionary entries show synonyms.

To summarize, we first randomly choose a POS tag and find a target token which has this tag in our utterance. Then we replace the target with a special $[$MASK$]$ token. We feed this utterance into BERT and obtain a list of candidates with their probabilities.

In case a sentence contains no token with the target POS, we use the more common \textit{noun} POS. We observe an example in the \textit{synonym adv.} row in Table \ref{tab:examples}.

\begin{table}
  \centering
	
	\begin{savenotes}
  \renewcommand{\arraystretch}{1.4}
  \setlength{\tabcolsep}{3pt}
    \scriptsize
    \centering
        \begin{tabular}{lr}
        \toprule
  \bf evaluation set & Example sentence \\
  \midrule
  Original & add \udensdot{tune} to \udensdash{sxsw fresh} playlist \\
  \midrule
  BOS Filler & \textbf{okay so} add \udensdot{tune} to \udensdash{sxsw fresh} playlist \\
  Pre-V. Filler & \textbf{like} add \udensdot{tune} to \udensdash{sxsw fresh} playlist \\
  Post-V. Filler & add \udensdot{tune} \textbf{actually} to \udensdash{sxsw fresh} playlist \\
  EOS Filler & add \udensdot{tune} to \udensdash{sxsw fresh} playlist \textbf{if you can} \\
  \midrule
  Synonym V. & \textbf{play} \udensdot{tune} to \udensdash{sxsw fresh} playlist \\
  Synonym Adj. & add \udensdot{tune} to \udensdash{sxsw \textbf{cool}} playlist \\
  Synonym Adv. & add \udensdot{\textbf{prior}} to \udensdash{sxsw fresh} playlist \\
  Synonym Any & \textbf{mix} \udensdot{tune} to \udensdash{sxsw fresh} playlist \\
  Synonym StopW & add \udensdot{tune} \textbf{the} \udensdash{sxsw fresh} playlist \\
  \midrule
  Speako & add \udensdot{\textbf{tua}} to \udensdash{sxsw fresh} playlist \\
    \bottomrule
        \end{tabular}
\caption{Processed variants of original utterances from the SNIPS corpus. The tokens labeled as \textit{music\_item} appear with a dotted underline and the tokens labeled as \textit{playlist} show a dashed underline. In SNIPS, the \textit{sxsw} token is part of a playlist name and an abbreviation of \textit{South by Southwest}.}
\label{tab:examples}
\end{savenotes}
\end{table}

\begin{figure}
    \centering
    \includegraphics[width=0.5\textwidth]{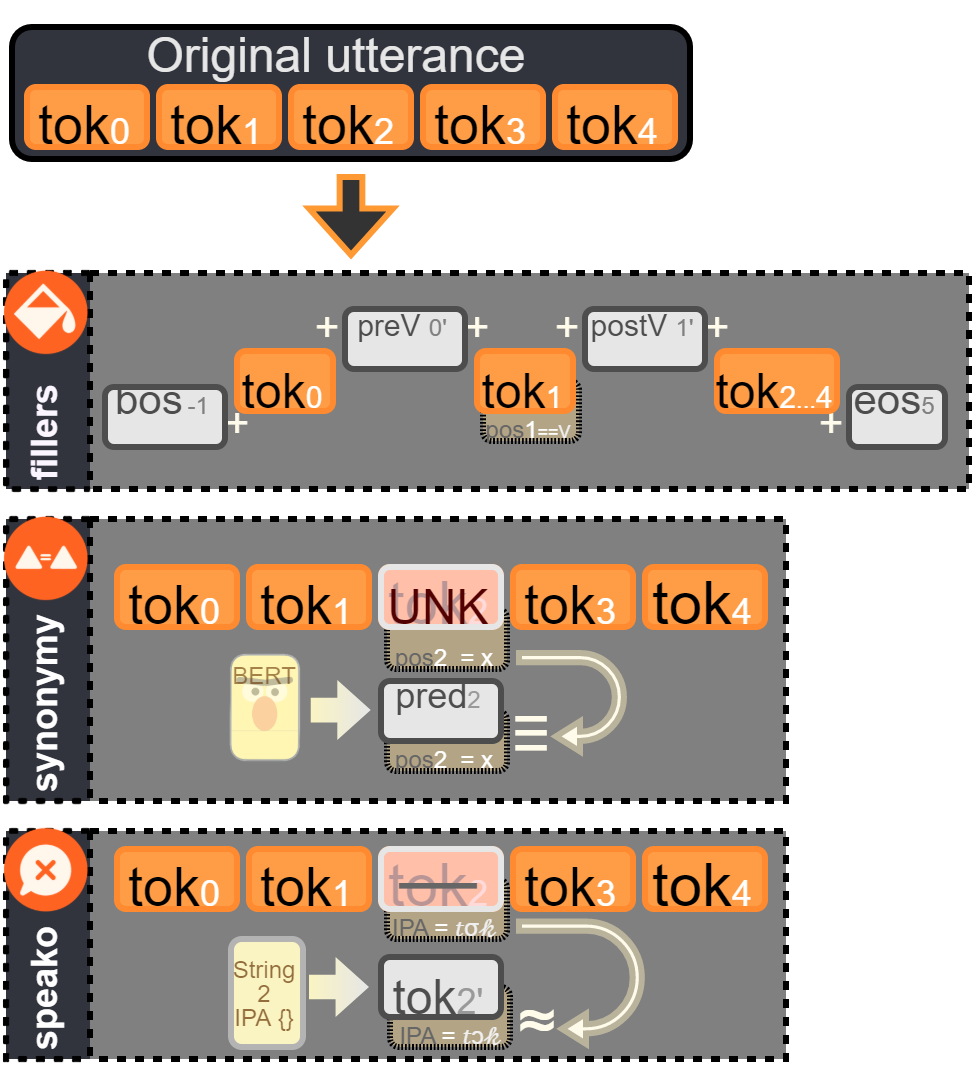}
    \caption{Overview of how each operator alters the original sentence, according to its type (filler, synonymy, speako).}
    \label{fig:pipeline}
\end{figure}

\begin{table}
  \centering
  \begin{savenotes}
\renewcommand{\arraystretch}{0.9}
\setlength{\tabcolsep}{3pt}
    \scriptsize
    \begin{tabular}{lll}
    \toprule
         \textbf{\makecell{Token in context}} &
         \textbf{\makecell{Wiktionary synonyms}} & \textbf{\makecell{ BERT candidates}} \\
         \midrule
         let me $\underset{\text{verb}}{\underline{\text{buy}}}$ it & \makecell{purchase,\\ accept, [...]} & \makecell{get, buy, present, make, \\ purchase, offer, give, sell,  [...]}\\
         \midrule
         is it $\underset{\text{adj}}{\underline{\text{large}}}$ ? & \makecell{giant, big, \\ huge, [...]} & \makecell{unusual, big, dangerous,\\ large, powerful, [...]}\\
         \midrule
         i said it $\underset{\text{adv}}{\underline{\text{quickly}}}$ & \makecell{rapidly,\\ fast} & \makecell{fast, well, strong, high,\\ good, deep, large, slow, [...]}\\
         \midrule
         give me $\underset{\text{noun}}{\underline{\text{freedom}}}$ & \makecell{liberty, \\ license, [...]} & \makecell{rights, property, freedom,\\ status, goods, liberty, [...]}\\
         \midrule
         i found $\underset{\text{stopword}}{\underline{\text{the}}}$ ball & \makecell{le} & \makecell{the, second, also, third,\\ their, still, a, our, 2nd, [...]}\\
    \bottomrule
    \end{tabular}
    \caption{Target words (underlined) of various POS and their synonyms taken from the crowd-sourced dictionary Wiktionary and candidates obtained using a pre-trained BERT language model.}
    \label{tab:synonym_examples}
\end{savenotes}
\end{table}

As we can see in Table \ref{tab:synonym_examples}, not all BERT candidates are suitable synonyms of the target token. We remove candidates that do not have the same POS of the target token. For a better performance, we place each candidate in the sentence before getting its POS. We have 5 different synonym operators based on different target POS: \textbf{verb}, \textbf{adjective}, \textbf{adverb},  \textbf{any} (at random between verb, adjective, adverb or noun), \textbf{stop-words} (grammatical and most common words).

\subsection{Speako}
\label{ssec:paronym}

Some words sound similar to others but have a different meaning altogether (e.g., \textit{decent} and \textit{descent}, \textit{this} and \textit{these}). This operator is based on the idea that anyone can make an error, but an efficient and robust model should be able to recover a minor mistake using the context. Thus, we introduce speakos (slip of the tongue, speech-to-text misinterpretation). These slips of the tongue appear commonly in oral human-to-human communication. According to some studies \cite{jaeger2004kids, stemberger1989speech} they represent between 48 and 67.4\% of all oral errors, depending on the type of speaker. Although we do not find similar studies for user-machine communication, we know this phenomenon is not exclusive to human-to human communication and we expect them to appear in a similar amount.

To implement the speako operator, we use a prepared dictionary of tokens appearing 1000+ times in the whole English Wikipedia\footnote{We empirically observed that removing all tokens that had a co-occurrence lower than 1000 eliminated most of the nonsensical strings and extreme misspellings and conserved most functional words.}. We convert each entry of the dictionary into its representation in International Phonetic Alphabet (IPA). We randomly select one token from the sentence, and also convert it to IPA. We then calculate the similarity between it and the dictionary's entries (using Levenshtein distance) and replace it with the closest candidate. For instance, the sentence \textit{let me \textbf{watch} \unessential{(/\textipa{wAtS}/)} a comedy video} could be transformed into \textit{let me \textbf{which} \unessential{(/\textipa{wItS}/)} a comedy video}).

Figure~\ref{fig:pipeline} shows how we alter the original utterance by the filler, synonymy and speako operators.

\section{Experimental Setup}
\label{sec:experiments}

\subsection{Data}

In our work, we use 3 popular open-source benchmarks \footnote{We do not consider datasets for other VA related tasks such as multi-intent detection (e.g., MixATIS and MixSNIPS \cite{Qin2020AGIFAA}) or multi-turn dialog (e.g., SGD dataset \cite{rastogi2020towards}).} which are summarized in Table \ref{tab:atis_SNIPS}:

\begin{description}
    \item[Airline Travel Information System (ATIS)]\footnote{CGNU General Public License, version 2} \cite{hemphill1990atis} introduced an NLU benchmark for the SF and ID tasks with 18 different intent labels, 127 slot labels and a vocabulary of 939 tokens. It contains annotated utterances corresponding to flight reservations, spoken dialogues and requests.
    \item[SNIPS]\footnote{Creative Commons Zero v1.0 Universal License} \cite{coucke2018SNIPS} proposed the SNIPS voice platform, from which a dataset of queries for the SF and ID tasks with 7 intent labels, 72 slot labels and a vocabulary of 12k tokens were extracted.
    \item[NLU-ED]\footnote{Creative Commons Attribution 4.0 International License} is a dataset of 25K human annotated utterances using the Amazon Mechanical Turk service \cite{liu2019benchmarking}. This NLU benchmark for the SF and ID tasks is comprised of 69 intent labels, 108 slot labels and a vocabulary of 7.9k tokens.
\end{description}

Following the common practice in the field \cite{hakkani2016multi, goo2018slot, qin2019stack, razumovskaia2021crossing, krishnan2021multilingual}), we report the performance of SF using the F1 score. Moreover, we propose an end-to-end accuracy (E2E) metric (sometimes referred in the literature as the sentence-level semantic accuracy \cite{qin2019stack}). This metric counts true positives when all the predicted labels (intent+slots) match the ground truth labels. This allows us to combine the SF and ID performance in a single more strict metric.

\begin{savenotes}
\begin{table}[th]
\renewcommand{\arraystretch}{0.9}
\setlength{\tabcolsep}{3pt}
    \scriptsize
    \centering
    \begin{tabular}{llrrr}
    \toprule
    \bf{Benchmark} & & \bf{Train} & \bf{Valid.} & \bf{Eval.} \\ 
    \midrule
    ATIS & Sent & 4 478 & 500 & 893 \\
     & Words & 50 497 & 5 703 & 9 164 \\
     & Voc & 867 & 463 & 448 \\
    \midrule
    SNIPS & Sent & 13 084 & 700 & 700 \\
     & Words & 117 700 & 6 384 & 6 354 \\
     & Voc & 11 418 & 1 571 & 1 624 \\
    \midrule
    NLU-ED & Sent & 20 628 & 2 544 & 2 544 \\
     & Words & 145 950 & 18 167 & 17 347 \\
     & Voc & 7 010 & 2 182 & 2 072 \\
    \bottomrule
    \end{tabular}
    \caption{Dataset size information of ATIS, SNIPS and NLU-ED benchmarks.}
    \label{tab:atis_SNIPS}
\end{table}
\end{savenotes}

Any dialog-based dataset extracted from real user situations has the potential of containing private and security sensitive information. This is the main cause for the relatively low amount of datasets for SF and ID. The benchmarks we mention are well known and cautiously cleaned (as presented in Section \ref{sec:methodo}). NATURE operators purposely avoid using any type of resource that would contain personal information. To the best of our knowledge, our work is not detrimental to people's safety, privacy, security, rights or to the environment in any way.

\subsection{Models}

We use two different state-of-the-art models: 
\begin{description}
    \item[Stack-Prop+BERT] \citep{qin2019stack} uses BERT as a token-level encoder that feeds into two different BiLSTMs, one per each task. The output of the SF BiLSTM is added to the ID BiLSTM input in order to produce a token-level intent prediction which is further averaged into a sentence-level prediction.
    \item[Bi-RNN] \citep{wang2018bi} uses two correlated BiLSTMs that cross-impact each other by accessing the other's hidden states and come to a joint prediction for SF and ID.
\end{description}

The pre-trained version of these models were not available\footnote{\url{https://github.com/LeePleased/StackPropagation-SLU} and \url{https://github.com/ray075hl/Bi-Model-Intent-And-Slot}}. For ATIS and SNIPS, we trained the models using the same hyperparameters proposed in the documentation by \cite{qin2019stack}\footnote{300 epochs, 0.001 learning rate, 0.4 dropout rate, 256 encoder hidden dimensions, 1024 attention hidden dimensions, 128 attention output dimensions, 256 word embedding dimensions for ATIS and 32 for SNIPS.} and \cite{wang2018bi}\footnote{500 epochs, max sentence length of 120, 0.001 learning rate, 0.2 dropout rate, 300 word embedding size, 200 LSTM hidden size}, respectively. For NLU-ED, we use the hyperparameters from SNIPS, as their size is comparable. The trained models obtained comparable results to their published counterpart (see in Appendix Section A). To train the models, we used 1 NVIDIA Tesla V100. It took between 3 and 71 hours to train the Stack-Prop+BERT model \citep{qin2019stack} (depending on the size of the benchmark), and between 68 and 130 hours to train the Bi-RNN model~\citep{wang2018bi}.

\subsection{Modified NATURE Evaluation Sets}
\label{sec:modified_test_sets}
Since the original evaluation sets only cover a limited set of patterns, we transform them by applying the NATURE patterns to obtain evaluation sets of the same size as the original ones. 
As previously illustrated, NATURE operators offer simple ways of perturbing utterances. In order to avoid rendering utterances unrecognizable from their original version, we only apply one operator at a time and only once in the sentence (e.g. we add 1 filler or synonymize 1 token or transform 1 token into its speako version). We design 2 NATURE experimental evaluation sets: \textit{Random} and \textit{Hard}. 
In the Random setting, for each utterance, we apply one operator at random and we repeat the random operator selection 10 times and calculate the mean score.

For the Hard setting, we use the popular BERT fine-tuning model~\cite{devlin2018bert}~\footnote{More specifically, JointBERT~\cite{chen2019bert} implemented at~\url{https://github.com/monologg/JointBERT}.} to filter-in the most challenging operators. For each evaluation utterance, we select the operator with the lowest confidence score (probability of the true class). In Table \ref{tab:hard_distrib} we show the operator composition (by percentage) of the Hard evaluation sets for each dataset.  

The Random evaluation set is meant to show how a random small change in the sentence can influence evaluation while the Hard evaluation set is meant to assess the lower-bound performance of how much the model depends on similar pattern sentences to obtain the correct prediction.

\begin{table}[th]
    \begin{savenotes}
    \setlength{\tabcolsep}{3pt}
        \scriptsize
        \centering
        \begin{tabular}{l|rrr}
        \toprule
            \bf Operator  & \bf ATIS &\bf SNIPS &\bf NLU-ED \\
            \midrule
            BOS Filler & 0.8 & 0.1 & 2.5 \\
            Pre-V. Filler & 6.0 & 3.7 & 16.0 \\
            Post-V. Filler & 1.9 & 8.6 & 5.1 \\
            EOS Filler & 9.0 & 52.3 & 8.3 \\
            Syn. V. & 25.6 & 5.4 & 16.3 \\
            Syn. Adj. & 29.2 & 15.0 & 23.4 \\
            Syn. Adv. & 11.8 & 5.6 & 10.2 \\
            Syn. Any & 5.3 & 1.1 & 4.8 \\
            Syn. StopW & 3.2 & 2.7 & 6.4  \\
            Speako & 7.2 & 5.4 & 6.9 \\
        
        \bottomrule
        \end{tabular}
        \caption{Composition (by percentage) of JointBERT-selected operators for the Hard experimental evaluation set.}
        \label{tab:hard_distrib}
    \end{savenotes}
\end{table}

\section{Results and Discussion}
\label{ssec:eval}

\subsection{Qualitative Evaluation} 
\label{ssec:valid}

Our assumption is that the operator-generated utterances share the same meaning and labeling as the original sentence. In order to measure this, we conducted a small but representative multiple-choice survey. We select 120 operator-perturbed utterances from the ATIS, SNIPS and NLU-ED benchmarks. We selected at random 40 utterances from each benchmark, making sure they were also evenly distributed between operators (12 utterances per operator). In addition to these, we cherry-picked 12 original utterances of high-quality that served as control. As we can see in the Survey Table in the Appendix Section C, the control scores stayed high and therefore, there was no reason to invalidate any participant's annotations.

14 participants (NLP and ML researchers, with no links to this work) volunteered to participate in this unpaid survey and consented verbally to the use of their data within the scope of this research. To avoid a decrease in annotation quality (due to fatigue), we split the participants in 2 groups of 7 members and divided the utterances in two sets (each with 60 operator-perturbed + 12 control utterances). We estimated the survey time to be 30-60 minutes, which was not far from the actual time (27-53 minutes).

For each utterance, we asked the participants to evaluate the intent and slot labels as \textit{reasonable} or \textit{unreasonable}.

\begin{table}[th]
    \begin{savenotes}
    \renewcommand{\arraystretch}{0.9}
    \setlength{\tabcolsep}{3pt}
        \centering
        \begin{tabular}{l|ll|ll}
        \toprule
        & \multicolumn{2}{c|}{Group 1} & \multicolumn{2}{c}{Group 2} \\
        & Experiment & Control & Experiment & Control \\
        \midrule
        Slot & 94.5 & 94.0 & 93.8 & 97.0 \\
        Intent & 89.0 & 97.6 & 85.9 & 97.5 \\
        \bottomrule
        \end{tabular}
        \caption{Survey results and statistics per group. All scores appear as percentages and indicate how the samples were perceived. A lower score indicates that more tokens and utterances have an \textit{unreasonable} label.}
        \label{tab:survey_result}
    \end{savenotes}
\end{table}

In Table \ref{tab:survey_result} we observe a sizable decrease on the experiment side for Intent, which can be partially explained by the disposition of some operators to perturb word types (such as verbs) that are highly associated with the intent classification. We also observe that the Slot labeling results are high and very close to the control scores. This indicates that (contrary to many DA strategies) the NATURE operators maintain a close-to-ground-truth slot labeling.

\begin{savenotes}
\begin{table*}[!htp]
\renewcommand{\arraystretch}{0.9}
\setlength{\tabcolsep}{3pt}
    \small
    \centering
    \begin{tabular}{l|lll|lll|lll||lll}
    \toprule
  \multirow{2}{*}{Evaluation Set}  & \multicolumn{3}{c|}{\textbf{ATIS}} & \multicolumn{3}{c|}{\textbf{SNIPS}} & \multicolumn{3}{c||}{\textbf{NLU-ED}} &  \multicolumn{3}{c}{\textbf{Avg.}} \\
  & \vtop{\hbox{\strut Slot}\hbox{\strut (F1)}} & \vtop{\hbox{\strut Intent}\hbox{\strut (Acc)}} & \vtop{\hbox{\strut E2E}\hbox{\strut (Acc)}} & \vtop{\hbox{\strut Slot}\hbox{\strut (F1)}} & \vtop{\hbox{\strut Intent}\hbox{\strut (Acc)}} &  \vtop{\hbox{\strut E2E}\hbox{\strut (Acc)}} & \vtop{\hbox{\strut Slot}\hbox{\strut (F1)}} & \vtop{\hbox{\strut Intent}\hbox{\strut (Acc)}} &  \vtop{\hbox{\strut E2E}\hbox{\strut (Acc)}} &  \vtop{\hbox{\strut Slot}\hbox{\strut (F1)}} & \vtop{\hbox{\strut Intent}\hbox{\strut (Acc)}} &  \vtop{\hbox{\strut E2E}\hbox{\strut (Acc)}}\\
  \midrule
  \multicolumn{13}{c}{Stack-Prop+BERT} \\
  \midrule
  Orig & 95.7 & 96.5 & 86.2 & 95.0 & 98.3 & 87.9 & 74.0 & 85.1 & 67.8 &  88.2 & 93.3 & 80.6 \\
  Rand & 91.3 & 95.0 & 66.5 & 83.4 & 96.1 & 53.8 & 67.4 & 76.1 & 56.8 &  80.7 & 89.1 & 59.0 \\
  Hard & 82.3 & 90.7 & 34.9 & 70.6 & 95.3 & 12.9 & 55.5 & 62.7 & 38.9 &  69.5 & 82.9 & 28.9 \\
  \midrule
Pre-V. Filler & 95.6 & 96.5 & 85.6 & 92.2 & 98.3 & 79.3 & 71.0 & 83.6 & 65.7 &  86.3 & 92.8 & 76.9 \\
  Syn. StopW & 93.0 & 94.8 & 76.5 & 89.7 & 96.7 & 74.3 & 70.2 & 78.9 & 60.2 &  84.3 & 90.1 & 70.3 \\
  BOS Filler & 95.6 & 96.2 & 85.8 & 86.5 & 97.1 & 54.9 & 72.5 & 80.8 & 63.9 &  84.9 & 91.4 & 68.2 \\
  Post-V. Filler & 94.0 & 96.5 & 80.3 & 84.8 & 98.0 & 57.1 & 68.0 & 84.1 & 63.6 &  82.3 & 92.9 & 67.0 \\
  Syn. V. & 90.1 & 95.3 & 63.6 & 88.4 & 95.1 & 66.7  & 68.5 & 74.2 & 56.5 &  82.3 & 88.2 & 62.3 \\
  Speako & 92.9 & 92.7 & 72.5 & 77.9 & 94.6 & 45.3  & 69.5 & 74.2 & 57.6 &  80.1 & 87.2 & 58.5 \\
  Syn. Any & 90.3 & 90.5 & 54.4 & 86.9 & 94.4 & 61.6 & 67.8 & 71.0 & 53.5 &  81.7 & 85.3 & 56.5 \\
  Syn. Adj. & \underline{84.7} & 92.7 & \underline{42.4} & 78.2 & 95.4 & 44.4  & \underline{60.2} & 69.7 & 47.2 &  74.4 & 85.9 & 44.7 \\
  Syn. Adv. & 88.2 & \underline{89.1} & 43.9 & 77.6 & \underline{94.3} & 41.9 & 61.6 & \underline{65.6} & \underline{45.4} &  75.8 & 83.0 & 43.7 \\
  EOS Filler & 88.9 & 96.3 & 54.1 & \underline{72.1} & 97.7 & \underline{13.1}  & 63.9 & 78.0 & 53.6 &  75.0 & 90.7 & 40.3 \\
  \midrule
  \multicolumn{13}{c}{Bi-RNN} \\
  \midrule
  Orig & 94.9 & 97.6 & 84.7 & 89.4 & 97.1 & 76.6 & 66.4 & 80.9 & 61.7 &  83.6 & 91.9 & 74.3 \\
  Rand & 89.9 & 94.3 & 61.8 & 75.6 & 94.1 & 39.0 & 60.6 & 70.8 & 50.1 &  75.4 & 86.4 & 50.3 \\
  Hard & 79.9 & 92.0 & 27.6 & 62.4 & 92.9 & 7.0 & 49.6 & 58.8 & 34.4 & 64.0 & 81.2 & 23.0 \\
  
  \midrule
  Pre-V. Filler & 94.7 & 97.3 & 82.2 & 84.6 & 96.4 & 60.0 & 63.3 & 80.1 & 59.3 &  80.9 & 91.3 & 67.2 \\
  Syn. StopW & 90.6 & 94.7 & 72.7 & 80.5 & 95.4 & 56.4  & 62.3 & 73.2 & 52.7 &  77.8 & 87.8 & 60.6 \\
  BOS Filler & \underline{80.7} & 96.7 & 82.6 & 80.9 & 96.7 & 38.4 & 65.8 & 78.8 & 59.6 &  75.8 & 90.7 & 60.2 \\
  Post-V. Filler & 93.8 & 96.9 & 80.3 & 77.9 & 96.6 & 37.4  & 62.6 & 79.3 & 56.6 &  78.1 & 90.9 & 58.1 \\
  Syn. V. & 87.6 & 95.9 & 56.6 & 79.5 & 92.1 & 50.6 & 61.3 & 70.5 & 50.7 &  76.1 & 86.2 & 52.6 \\
  Speako & 91.8 & 90.3 & 68.1 & 70.1 & \underline{90.1} & 33.6  & 61.5 & 69.8 & 51.0 &  74.5 & 83.4 & 50.9 \\
  Syn. Any & 89.2 & 90.4 & 52.6 & 77.8 & 91.4 & 40.6 & 62.0 & 67.3 & 49.1 &  76.3 & 83.0 & 47.4 \\
  Syn. Adj. & 81.7 & 94.2 & \underline{34.4} & 71.7 & 93.9 & 34.9 & \underline{54.3} & 65.5 & 42.1 &  69.2 & 84.5 & 37.1 \\
  Syn. Adv. & 87.2 & \underline{85.1} & 38.4 & 69.9 & 92.1 & 29.0 & 54.7 & \underline{61.4} & \underline{40.3} &  70.6 & 79.5 & 35.9 \\
  EOS Filler & 88.9 & 96.8 & 52.2 & \underline{64.1} & 94.1 & \underline{5.9} & 56.4 & 65.8 & 42.0 &  69.8 & 85.6 & 33.4 \\
  \bottomrule
\end{tabular}
\caption{SF, ID and E2E performances of BERT and RNN based models trained on ATIS, SNIPS, and NLU-ED and evaluated on their original and NATURE-perturbed evaluation sets. We show results on \textit{per-operator} as well as on Random and Hard evaluation sets. Furthermore, we report the unweighted average score on the 3 benchmark we considered. The lowest scores in each column appear underlined.}
\label{tab:res_main}
\end{table*}
\end{savenotes}

\begin{savenotes}
\begin{table*}[!htp]
\renewcommand{\arraystretch}{0.9}
\setlength{\tabcolsep}{3pt}
    \small
    \centering
    \begin{tabular}{l|lll|lll|lll||lll}
    \toprule
  \multirow{2}{*}{Evaluation Set}  & \multicolumn{3}{c|}{\textbf{ATIS}} & \multicolumn{3}{c|}{\textbf{SNIPS}} & \multicolumn{3}{c||}{\textbf{NLU-ED}} &  \multicolumn{3}{c}{\textbf{Avg.}} \\
  & \vtop{\hbox{\strut Slot}\hbox{\strut (F1)}} & \vtop{\hbox{\strut Intent}\hbox{\strut (Acc)}} & \vtop{\hbox{\strut E2E}\hbox{\strut (Acc)}} & \vtop{\hbox{\strut Slot}\hbox{\strut (F1)}} & \vtop{\hbox{\strut Intent}\hbox{\strut (Acc)}} &  \vtop{\hbox{\strut E2E}\hbox{\strut (Acc)}} & \vtop{\hbox{\strut Slot}\hbox{\strut (F1)}} & \vtop{\hbox{\strut Intent}\hbox{\strut (Acc)}} &  \vtop{\hbox{\strut E2E}\hbox{\strut (Acc)}} &  \vtop{\hbox{\strut Slot}\hbox{\strut (F1)}} & \vtop{\hbox{\strut Intent}\hbox{\strut (Acc)}} &  \vtop{\hbox{\strut E2E}\hbox{\strut (Acc)}}\\
  \midrule
  \multicolumn{13}{c}{Stack-Prop+BERT} \\
  \midrule
  Orig & 95.7 & 96.5 & 86.2 & 95.0 & 98.3 & 87.9 & 74.0 & 85.1 & 67.8 &  88.2 & 93.3 & 80.6 \\
  \midrule
  Checklist Contract. & 95.6 & 96.6 & 85.8 & 94.6 & 98.2 & 86.8 & 73.8 & 84.6 & 67.4 & 88.0 & 93.1 & 80.0 \\
  Checklist NER & 94.7 & 96.5 & 84.6 & 92.9 & 98.2 & 83.0 & 73.7 & 85.1 & 67.6 & 87.1 & 93.3 & 78.4 \\
  Checklist Typo & \underline{85.1} & \underline{92.2} & 51.0 & 78.3 & \underline{95.4} & 51.7 & 57.1 & 70.8 & 46.9 & 73.5 & 86.1 & 49.9 \\
  Checklist Punct. & 85.2 & 96.6 & \underline{42.8} & \underline{71.7} & 97.7 & \underline{20.4} & \underline{55.4} & \underline{26.3} & \underline{16.7} & 70.8 & 73.5 & 26.6 \\
  \midrule
  \multicolumn{13}{c}{Bi-RNN} \\
  \midrule
  Orig & 94.9 & 97.6 & 84.7 & 89.4 & 97.1 & 76.6 & 66.4 & 80.9 & 61.7 &  83.6 & 91.9 & 74.3 \\
  \midrule
  Checklist Contract. & 94.8 & 97.5 & 84.3 & 88.9 & 96.5 & 75.8 & 67.0 & 81.1 & 61.0 & 83.6 & 91.7 & 73.7 \\
  Checklist NER & 93.8 & 97.5 & 83.0 & 89.5 & 96.7 & 78.2 & 67.6 & 81.7 & 61.5 & 83.6 & 92.0 & 74.2 \\
  Checklist Typo & \underline{81.6} & \underline{}{92.1} & 43.2 & \underline{70.1} & \underline{92.7} & 37.8 & \underline{49.5} & 66.4 & 41.6 & 67.1 & 83.7 & 40.9  \\
  Checklist Punct. & 87.4 & 96.7 & \underline{40.7} & 71.0 & 96.2 & \underline{20.7} & 50.0 & \underline{41.4} & \underline{22.4} & 69.5 & 78.1 & 27.9 \\
  \bottomrule
\end{tabular}
\caption{SF, ID and E2E performances of BERT and RNN based models trained on ATIS, SNIPS, and NLU-ED and evaluated on their original and CHECKLIST-perturbed evaluation sets. The lowest scores in each column appear underlined.}
\label{tab:res_check}
\end{table*}
\end{savenotes}

\subsection{Quantitative Evaluation}
\label{ssec:quant_eval}

Table~\ref{tab:res_main} shows the performances of the Stack-Prop+BERT and Bi-RNN models trained on the original train data of ATIS, SNIPS and NLU-ED benchmarks. Models are evaluated on the Original, Random and Hard evaluation sets. We also show the scores on 10 evaluation sets, each perturbed with a single NATURE operator, where one operator is applied once to each utterance of the evaluation set. In Table~\ref{tab:res_main}, for each benchmark, we report the F1 and accuracy on the SF and ID tasks respectively, and the E2E metric. Furthermore, we report the unweighted average (Avg. column) of the aforementioned scores on the three benchmarks. The perturbed evaluation set results are sorted in descending order according to the averaged E2E metric. We notice that Stack-Prop+BERT outperforms Bi-RNN not only on original, but also on all evaluation set variants. More precisely, we observe a gap of 6.3\%, 8.7\% and 5.9\% on the \textit{Avg.} E2E metric on the Orig, Random and Hard evaluation sets.

First, we observe a noticeable lowering in the scores on Random, and quite a radical change on Hard. We consider the possibility that the Hard evaluation set incorporates more noise than the Random evaluation sets, and this could be the cause of this low score. However, depending on the benchmark, the sharpest operators are not always the ones expected to be most disruptive. Yet, the decrease in score is extreme across all benchmarks and for both models.

As mentioned earlier, fillers contribute little to the semantics of an utterance and should not be disruptive for the model. The speako operator is more disruptive semantically, specially for the cases where the original token cannot be deduced from the context and the perturbed token. We expect the synonym operators to be the most disruptive of the three since we modify the semantic value of a whole word at a time. 
In Table \ref{tab:res_main} we observe that the model handles most filler operators reasonably well, however, we are surprised to see the scores drop considerably for the EOS. As shown on Table~\ref{tab:res_main}, the EOS operator drops the E2E accuracy of both models by about 40\% on average across all benchmarks. This suggests some syntax-level pattern dependence where the models use the position of the tokens to achieve the correct predictions. The synonym operators, specially the adverb and adjective, greatly deteriorate the performances. This decrease in score shines a light on the importance of the token-level pattern, signaling that the models are using certain adjectives and adverbs to make their predictions. Since, in the benchmarks, adjectives and adverbs are much less diverse than the nouns and verbs, we infer that the models are using these words as prediction clues. The speako operator scores suggest a good capacity of the models to overcome these variants and generalize using the remaining context.

Interestingly, we notice that the drop of performances is highly strong on the E2E metric. For instance, using the Stack-Prop+BERT model on the ATIS evaluation set, perturbed with the \texttt{EOS filler} operator, we observe a 0.3\% and 6.8\% drop on SF and ID respectively but a 32.1\% drop on E2E. 
We use the E2E metric since it is more representative of the whole frame accuracy of real world scenarios ~\cite{goo2018slot, qin2019stack}, where a VA can only execute the expected command if the intent and all slots are correctly predicted. A more concise illustration of Table~\ref{tab:res_main}~'s results is shown on Figure 2 in the Appendix Section D.

We also apply a general perturbation method, the Checklist tool \cite{ribeiro2020beyond}\footnote{\url{https://github.com/marcotcr/checklist}} to evaluate model performance on the three benchmarks. Although not designed for spoken language, some of the operators are useful to diagnose the problem of over-fitting to spurious patterns and correlations. We use 4 that are suitable for most sentences: the \textbf{punctuation} operator removes or adds a final punctuation according to its presence or absence in the text, the \textbf{typo} operator swaps random characters with one of its neighbours, the \textbf{contraction} operator replaces contracted words with their non-contracted version or vice-versa (e.g., \textit{don't} $\rightarrow$ \textit{do not}, \textit{cannot} $\rightarrow$ \textit{can't}), the \textbf{NER} operator detects and replaces first names, locations and numbers with other named entities of the same type.  Table \ref{tab:res_check} shows the results and we observe that the punctuation operator can reduce the E2E accuracy of both models by more than 45\% on average across all benchmarks. This supports the results we observed on NATURE.

\section{Conclusions}
\label{sec:conclusions}
Neural Network models have a black-box architecture that makes it hard to discern when they correctly generalize over the input and when they resort to heuristic features that correlate to the expected output. 

We present the NATURE operators, apply them to evaluation sets of standard SF and ID benchmarks and observe a significant drop of the state-of-the-art model scores. The different operators in our framework help discern what surface patterns the model is exploiting.

These results should hopefully encourage the development of better, more challenging benchmarks and the search for more robust models, capable of handling more realistic, fitting and spoken-language oriented utterances.

For future work, we wish to expand the NATURE operators to include speech impediments (such as lisp, stutter and dysarthria), extend the operators to be multi-lingual and work on multi-turn dialogue and multi-intent detection tasks.

\section{Acknowledgments}
\label{sec:acknow}
We would like to thank the team at Mindspore\footnote{\url{https://www.mindspore.cn/}}, a new deep learning computing framework, for partial support on this work. Moreover, we want to thank Prasanna Parthasarathi for his valuable feedback and suggestions and the survey volunteers for their time and participation.

\bibliography{natbib}

\begin{thebibliography}{55}
\providecommand{\natexlab}[1]{#1}
\providecommand{\url}[1]{\texttt{#1}}
\expandafter\ifx\csname urlstyle\endcsname\relax
  \providecommand{\doi}[1]{doi: #1}\else
  \providecommand{\doi}{doi: \begingroup \urlstyle{rm}\Url}\fi

\bibitem[Balasubramanian et~al.(2020)Balasubramanian, Jain, Jindal, Awasthi,
  and Sarawagi]{balasubramanian2020s}
Sriram Balasubramanian, Naman Jain, Gaurav Jindal, Abhijeet Awasthi, and Sunita
  Sarawagi.
\newblock What's in a name? are bert named entity representations just as good
  for any other name?
\newblock \emph{arXiv preprint arXiv:2007.06897}, 2020.

\bibitem[Belinkov and Bisk(2017)]{belinkov2017synthetic}
Yonatan Belinkov and Yonatan Bisk.
\newblock Synthetic and natural noise both break neural machine translation.
\newblock \emph{arXiv preprint arXiv:1711.02173}, 2017.

\bibitem[Bengio and Heigold(2014)]{bengio2014word}
Samy Bengio and Georg Heigold.
\newblock Word embeddings for speech recognition.
\newblock 2014.

\bibitem[Chen et~al.(2019)Chen, Zhuo, and Wang]{chen2019bert}
Qian Chen, Zhu Zhuo, and Wen Wang.
\newblock Bert for joint intent classification and slot filling.
\newblock \emph{arXiv preprint arXiv:1902.10909}, 2019.

\bibitem[Coucke et~al.(2018)Coucke, Saade, Ball, Bluche, Caulier, Leroy,
  Doumouro, Gisselbrecht, Caltagirone, Lavril, et~al.]{coucke2018SNIPS}
Alice Coucke, Alaa Saade, Adrien Ball, Th{\'e}odore Bluche, Alexandre Caulier,
  David Leroy, Cl{\'e}ment Doumouro, Thibault Gisselbrecht, Francesco
  Caltagirone, Thibaut Lavril, et~al.
\newblock Snips voice platform: an embedded spoken language understanding
  system for private-by-design voice interfaces.
\newblock \emph{arXiv preprint arXiv:1805.10190}, 2018.

\bibitem[Dai and Adel(2020)]{dai2020analysis}
Xiang Dai and Heike Adel.
\newblock An analysis of simple data augmentation for named entity recognition.
\newblock In \emph{Proceedings of the 28th International Conference on
  Computational Linguistics}, pages 3861--3867, 2020.

\bibitem[Dayanik and Pad{\'o}(2020)]{dayanik2020masking}
Erenay Dayanik and Sebastian Pad{\'o}.
\newblock Masking actor information leads to fairer political claims detection.
\newblock In \emph{Proceedings of the 58th Annual Meeting of the Association
  for Computational Linguistics}, pages 4385--4391, 2020.

\bibitem[Devlin et~al.(2018)Devlin, Chang, Lee, and Toutanova]{devlin2018bert}
Jacob Devlin, Ming-Wei Chang, Kenton Lee, and Kristina Toutanova.
\newblock Bert: Pre-training of deep bidirectional transformers for language
  understanding.
\newblock \emph{arXiv preprint arXiv:1810.04805}, 2018.

\bibitem[Ebrahimi et~al.(2017)Ebrahimi, Rao, Lowd, and
  Dou]{ebrahimi2017hotflip}
Javid Ebrahimi, Anyi Rao, Daniel Lowd, and Dejing Dou.
\newblock Hotflip: White-box adversarial examples for text classification.
\newblock \emph{arXiv preprint arXiv:1712.06751}, 2017.

\bibitem[Gardner et~al.(2020)Gardner, Artzi, Basmova, Berant, Bogin, Chen,
  Dasigi, Dua, Elazar, Gottumukkala, et~al.]{gardner2020evaluating}
Matt Gardner, Yoav Artzi, Victoria Basmova, Jonathan Berant, Ben Bogin, Sihao
  Chen, Pradeep Dasigi, Dheeru Dua, Yanai Elazar, Ananth Gottumukkala, et~al.
\newblock Evaluating models' local decision boundaries via contrast sets.
\newblock \emph{arXiv preprint arXiv:2004.02709}, 2020.

\bibitem[Garg and Ramakrishnan()]{gargbae}
Siddhant Garg and Goutham Ramakrishnan.
\newblock Bae: Bert-based adversarial examples for text classification.

\bibitem[Ghaddar et~al.(2021)Ghaddar, Langlais, Rashid, and
  Rezagholizadeh]{ghaddar2021context}
Abbas Ghaddar, Philippe Langlais, Ahmad Rashid, and Mehdi Rezagholizadeh.
\newblock Context-aware adversarial training for name regularity bias in named
  entity recognition.
\newblock \emph{Transactions of the Association for Computational Linguistics},
  9:\penalty0 586--604, 2021.

\bibitem[Godfrey et~al.(1992)Godfrey, Holliman, and
  McDaniel]{godfrey1992switchb}
John Godfrey, Edward Holliman, and Jane McDaniel.
\newblock Switchboard: Telephone speech corpus for research and development.
\newblock In \emph{Proceedings of the 1992 IEEE International Conference on
  Acoustics, Speech and Signal Processing (ICASSP)}, volume~1, pages 517--520.
  IEEE, 1992.

\bibitem[Goel et~al.(2021)Goel, Rajani, Vig, Tan, Wu, Zheng, Xiong, Bansal, and
  R{\'e}]{goel2021robustness}
Karan Goel, Nazneen Rajani, Jesse Vig, Samson Tan, Jason Wu, Stephan Zheng,
  Caiming Xiong, Mohit Bansal, and Christopher R{\'e}.
\newblock Robustness gym: Unifying the nlp evaluation landscape.
\newblock \emph{arXiv preprint arXiv:2101.04840}, 2021.

\bibitem[Goo et~al.(2018)Goo, Gao, Hsu, Huo, Chen, Hsu, and Chen]{goo2018slot}
Chih-Wen Goo, Guang Gao, Yun-Kai Hsu, Chih-Li Huo, Tsung-Chieh Chen, Keng-Wei
  Hsu, and Yun-Nung Chen.
\newblock Slot-gated modeling for joint slot filling and intent prediction.
\newblock In \emph{Proceedings of the 2018 Conference of the North American
  Chapter of the Association for Computational Linguistics: Human Language
  Technologies, Volume 2 (Short Papers)}, pages 753--757, 2018.

\bibitem[Gopalakrishnan et~al.(2020)Gopalakrishnan, Hedayatnia, Wang, Liu, and
  Hakkani-Tur]{gopalakrishnan2020neural}
Karthik Gopalakrishnan, Behnam Hedayatnia, Longshaokan Wang, Yang Liu, and
  Dilek Hakkani-Tur.
\newblock Are neural open-domain dialog systems robust to speech recognition
  errors in the dialog history? an empirical study.
\newblock \emph{arXiv preprint arXiv:2008.07683}, 2020.

\bibitem[Gururangan et~al.(2018)Gururangan, Swayamdipta, Levy, Schwartz,
  Bowman, and Smith]{gururangan2018annotation}
Suchin Gururangan, Swabha Swayamdipta, Omer Levy, Roy Schwartz, Samuel~R
  Bowman, and Noah~A Smith.
\newblock Annotation artifacts in natural language inference data.
\newblock \emph{arXiv preprint arXiv:1803.02324}, 2018.

\bibitem[Hakkani-T{\"u}r et~al.(2016)Hakkani-T{\"u}r, T{\"u}r, Celikyilmaz,
  Chen, Gao, Deng, and Wang]{hakkani2016multi}
Dilek Hakkani-T{\"u}r, G{\"o}khan T{\"u}r, Asli Celikyilmaz, Yun-Nung Chen,
  Jianfeng Gao, Li~Deng, and Ye-Yi Wang.
\newblock Multi-domain joint semantic frame parsing using bi-directional
  rnn-lstm.
\newblock In \emph{Interspeech}, pages 715--719, 2016.

\bibitem[Hemphill et~al.(1990)Hemphill, Godfrey, and
  Doddington]{hemphill1990atis}
Charles~T Hemphill, John~J Godfrey, and George~R Doddington.
\newblock The atis spoken language systems pilot corpus.
\newblock In \emph{Speech and Natural Language: Proceedings of a Workshop Held
  at Hidden Valley, Pennsylvania, June 24-27, 1990}, 1990.

\bibitem[Jaeger(2004)]{jaeger2004kids}
Jeri~J Jaeger.
\newblock \emph{Kids' slips: What young children's slips of the tongue reveal
  about language development}.
\newblock Psychology Press, 2004.

\bibitem[Jia and Liang(2017)]{jia2017adversarial}
Robin Jia and Percy Liang.
\newblock Adversarial examples for evaluating reading comprehension systems.
\newblock \emph{arXiv preprint arXiv:1707.07328}, 2017.

\bibitem[Jin et~al.(2019)Jin, Jin, Zhou, and Szolovits]{jin2019bert}
Di~Jin, Zhijing Jin, Joey~Tianyi Zhou, and Peter Szolovits.
\newblock Is bert really robust? natural language attack on text classification
  and entailment.
\newblock \emph{arXiv preprint arXiv:1907.11932}, 2019.

\bibitem[Jin et~al.(2020)Jin, Jin, Zhou, and Szolovits]{jin2020bert}
Di~Jin, Zhijing Jin, Joey~Tianyi Zhou, and Peter Szolovits.
\newblock Is bert really robust? a strong baseline for natural language attack
  on text classification and entailment.
\newblock In \emph{Proceedings of the AAAI conference on artificial
  intelligence}, volume~34, pages 8018--8025, 2020.

\bibitem[Kaushik et~al.(2019)Kaushik, Hovy, and Lipton]{kaushik2019learning}
Divyansh Kaushik, Eduard Hovy, and Zachary~C Lipton.
\newblock Learning the difference that makes a difference with
  counterfactually-augmented data.
\newblock \emph{arXiv preprint arXiv:1909.12434}, 2019.

\bibitem[Kiela et~al.(2021)Kiela, Bartolo, Nie, Kaushik, Geiger, Wu, Vidgen,
  Prasad, Singh, Ringshia, et~al.]{kiela2021dynabench}
Douwe Kiela, Max Bartolo, Yixin Nie, Divyansh Kaushik, Atticus Geiger,
  Zhengxuan Wu, Bertie Vidgen, Grusha Prasad, Amanpreet Singh, Pratik Ringshia,
  et~al.
\newblock Dynabench: Rethinking benchmarking in nlp.
\newblock \emph{arXiv preprint arXiv:2104.14337}, 2021.

\bibitem[Krishnan et~al.(2021)Krishnan, Anastasopoulos, Purohit, and
  Rangwala]{krishnan2021multilingual}
Jitin Krishnan, Antonios Anastasopoulos, Hemant Purohit, and Huzefa Rangwala.
\newblock Multilingual code-switching for zero-shot cross-lingual intent
  prediction and slot filling.
\newblock \emph{arXiv preprint arXiv:2103.07792}, 2021.

\bibitem[Li et~al.(2020)Li, Yavuz, Hashimoto, Li, Niu, Rajani, Yan, Zhou, and
  Xiong]{li2020coco}
Shiyang Li, Semih Yavuz, Kazuma Hashimoto, Jia Li, Tong Niu, Nazneen Rajani,
  Xifeng Yan, Yingbo Zhou, and Caiming Xiong.
\newblock Coco: Controllable counterfactuals for evaluating dialogue state
  trackers.
\newblock \emph{arXiv preprint arXiv:2010.12850}, 2020.

\bibitem[Li et~al.(2018)Li, Xue, Chen, Liu, Feng, and Liu]{li2018improving}
Xiang Li, Haiyang Xue, Wei Chen, Yang Liu, Yang Feng, and Qun Liu.
\newblock Improving the robustness of speech translation.
\newblock \emph{arXiv preprint arXiv:1811.00728}, 2018.

\bibitem[Lin et~al.(2020)Lin, Lu, Tang, Han, Sun, Wei, and
  Yuan]{lin2020rigorous}
Hongyu Lin, Yaojie Lu, Jialong Tang, Xianpei Han, Le~Sun, Zhicheng Wei, and
  Nicholas~Jing Yuan.
\newblock A rigorous study on named entity recognition: Can fine-tuning
  pretrained model lead to the promised land?
\newblock \emph{arXiv preprint arXiv:2004.12126}, 2020.

\bibitem[Lin et~al.(2012)Lin, Etzioni, et~al.]{lin2012entity}
Thomas Lin, Oren Etzioni, et~al.
\newblock Entity linking at web scale.
\newblock In \emph{Proceedings of the Joint Workshop on Automatic Knowledge
  Base Construction and Web-scale Knowledge Extraction (AKBC-WEKEX)}, pages
  84--88, 2012.

\bibitem[Liu et~al.(2019)Liu, Eshghi, Swietojanski, and
  Rieser]{liu2019benchmarking}
Xingkun Liu, Arash Eshghi, Pawel Swietojanski, and Verena Rieser.
\newblock Benchmarking natural language understanding services for building
  conversational agents.
\newblock \emph{arXiv preprint arXiv:1903.05566}, 2019.

\bibitem[Louvan and Magnini(2020)]{louvan2020simple}
Samuel Louvan and Bernardo Magnini.
\newblock Simple is better! lightweight data augmentation for low resource slot
  filling and intent classification.
\newblock \emph{arXiv preprint arXiv:2009.03695}, 2020.

\bibitem[McCoy et~al.(2019)McCoy, Pavlick, and Linzen]{mccoy2019right}
R~Thomas McCoy, Ellie Pavlick, and Tal Linzen.
\newblock Right for the wrong reasons: Diagnosing syntactic heuristics in
  natural language inference.
\newblock \emph{arXiv preprint arXiv:1902.01007}, 2019.

\bibitem[McCrae et~al.(2020)McCrae, Rademaker, Rudnicka, and
  Bond]{mccrae2020english}
John~Philip McCrae, Alexandre Rademaker, Ewa Rudnicka, and Francis Bond.
\newblock English wordnet 2020: improving and extending a wordnet for english
  using an open-source methodology.
\newblock In \emph{Proceedings of the LREC 2020 Workshop on Multimodal Wordnets
  (MMW2020)}, pages 14--19, 2020.

\bibitem[Min et~al.(2020)Min, McCoy, Das, Pitler, and Linzen]{min2020syntactic}
Junghyun Min, R~Thomas McCoy, Dipanjan Das, Emily Pitler, and Tal Linzen.
\newblock Syntactic data augmentation increases robustness to inference
  heuristics.
\newblock In \emph{Proceedings of the 58th Annual Meeting of the Association
  for Computational Linguistics}, pages 2339--2352, 2020.

\bibitem[Moosavi et~al.(2020)Moosavi, de~Boer, Utama, and
  Gurevych]{moosavi2020improving}
Nafise~Sadat Moosavi, Marcel de~Boer, Prasetya~Ajie Utama, and Iryna Gurevych.
\newblock Improving robustness by augmenting training sentences with
  predicate-argument structures.
\newblock \emph{arXiv preprint arXiv:2010.12510}, 2020.

\bibitem[Naik et~al.(2018)Naik, Ravichander, Sadeh, Rose, and
  Neubig]{naik2018stress}
Aakanksha Naik, Abhilasha Ravichander, Norman Sadeh, Carolyn Rose, and Graham
  Neubig.
\newblock Stress test evaluation for natural language inference.
\newblock \emph{arXiv preprint arXiv:1806.00692}, 2018.

\bibitem[Pad{\'o} et~al.(2019)Pad{\'o}, Blessing, Blokker, Dayanik, Haunss, and
  Kuhn]{pado2019sides}
Sebastian Pad{\'o}, Andr{\'e} Blessing, Nico Blokker, Erenay Dayanik, Sebastian
  Haunss, and Jonas Kuhn.
\newblock Who sides with whom? towards computational construction of discourse
  networks for political debates.
\newblock In \emph{Proceedings of the 57th Annual Meeting of the Association
  for Computational Linguistics}, pages 2841--2847, 2019.

\bibitem[Poliak et~al.(2018)Poliak, Naradowsky, Haldar, Rudinger, and
  Van~Durme]{poliak2018hypothesis}
Adam Poliak, Jason Naradowsky, Aparajita Haldar, Rachel Rudinger, and Benjamin
  Van~Durme.
\newblock Hypothesis only baselines in natural language inference.
\newblock \emph{arXiv preprint arXiv:1805.01042}, 2018.

\bibitem[Qin et~al.(2019)Qin, Che, Li, Wen, and Liu]{qin2019stack}
Libo Qin, Wanxiang Che, Yangming Li, Haoyang Wen, and Ting Liu.
\newblock A stack-propagation framework with token-level intent detection for
  spoken language understanding.
\newblock \emph{arXiv preprint arXiv:1909.02188}, 2019.

\bibitem[Qin et~al.(2020)Qin, Xu, Che, and Liu]{Qin2020AGIFAA}
Libo Qin, Xiao Xu, Wanxiang Che, and Ting Liu.
\newblock Agif: An adaptive graph-interactive framework for joint multiple
  intent detection and slot filling.
\newblock \emph{arXiv: Computation and Language}, 2020.

\bibitem[Rastogi et~al.(2020)Rastogi, Zang, Sunkara, Gupta, and
  Khaitan]{rastogi2020towards}
Abhinav Rastogi, Xiaoxue Zang, Srinivas Sunkara, Raghav Gupta, and Pranav
  Khaitan.
\newblock Towards scalable multi-domain conversational agents: The
  schema-guided dialogue dataset.
\newblock In \emph{Proceedings of the AAAI Conference on Artificial
  Intelligence}, volume~34, pages 8689--8696, 2020.

\bibitem[Razumovskaia et~al.(2021)Razumovskaia, Glava{\v{s}}, Majewska,
  Korhonen, and Vuli{\'c}]{razumovskaia2021crossing}
Evgeniia Razumovskaia, Goran Glava{\v{s}}, Olga Majewska, Anna Korhonen, and
  Ivan Vuli{\'c}.
\newblock Crossing the conversational chasm: A primer on multilingual
  task-oriented dialogue systems.
\newblock \emph{arXiv preprint arXiv:2104.08570}, 2021.

\bibitem[Ribeiro et~al.(2020)Ribeiro, Wu, Guestrin, and
  Singh]{ribeiro2020beyond}
Marco~Tulio Ribeiro, Tongshuang Wu, Carlos Guestrin, and Sameer Singh.
\newblock Beyond accuracy: Behavioral testing of nlp models with checklist.
\newblock \emph{arXiv preprint arXiv:2005.04118}, 2020.

\bibitem[Shriberg(2001)]{shriberg2001errrr}
Elizabeth Shriberg.
\newblock To ‘errrr’ is human: Ecology and acoustics of speech
  disfluencies.
\newblock \emph{Journal of the International Phonetic Association},
  31:\penalty0 153--169, 2001.

\bibitem[Simonnet et~al.(2018)Simonnet, Ghannay, Camelin, and
  Est{\`e}ve]{simonnet2018simulating}
Edwin Simonnet, Sahar Ghannay, Nathalie Camelin, and Yannick Est{\`e}ve.
\newblock Simulating asr errors for training slu systems.
\newblock In \emph{Proceedings of the Eleventh International Conference on
  Language Resources and Evaluation (LREC 2018)}, 2018.

\bibitem[Stemberger(1989)]{stemberger1989speech}
Joseph~Paul Stemberger.
\newblock Speech errors in early child language production.
\newblock \emph{Journal of Memory and Language}, 28\penalty0 (2):\penalty0
  164--188, 1989.

\bibitem[Tsvetkov et~al.(2014)Tsvetkov, Metze, and
  Dyer]{tsvetkov2014augmenting}
Yulia Tsvetkov, Florian Metze, and Chris Dyer.
\newblock Augmenting translation models with simulated acoustic confusions for
  improved spoken language translation.
\newblock In \emph{Proceedings of the 14th Conference of the European Chapter
  of the Association for Computational Linguistics}, pages 616--625, 2014.

\bibitem[Wang et~al.(2018)Wang, Shen, and Jin]{wang2018bi}
Yu~Wang, Yilin Shen, and Hongxia Jin.
\newblock A bi-model based rnn semantic frame parsing model for intent
  detection and slot filling.
\newblock \emph{arXiv preprint arXiv:1812.10235}, 2018.

\bibitem[Yamada et~al.(2020)Yamada, Asai, Shindo, Takeda, and
  Matsumoto]{yamada2020luke}
Ikuya Yamada, Akari Asai, Hiroyuki Shindo, Hideaki Takeda, and Yuji Matsumoto.
\newblock Luke: deep contextualized entity representations with entity-aware
  self-attention.
\newblock \emph{arXiv preprint arXiv:2010.01057}, 2020.

\bibitem[Yang et~al.(2019)Yang, Zhang, Tar, and Baldridge]{yang2019paws}
Yinfei Yang, Yuan Zhang, Chris Tar, and Jason Baldridge.
\newblock Paws-x: A cross-lingual adversarial dataset for paraphrase
  identification.
\newblock \emph{arXiv preprint arXiv:1908.11828}, 2019.

\bibitem[Yin et~al.(2020)Yin, Long, Meng, and Chang]{yin2020robustness}
Fan Yin, Quanyu Long, Tao Meng, and Kai-Wei Chang.
\newblock On the robustness of language encoders against grammatical errors.
\newblock \emph{arXiv preprint arXiv:2005.05683}, 2020.

\bibitem[Zellers et~al.(2018)Zellers, Bisk, Schwartz, and
  Choi]{zellers2018swag}
Rowan Zellers, Yonatan Bisk, Roy Schwartz, and Yejin Choi.
\newblock Swag: A large-scale adversarial dataset for grounded commonsense
  inference.
\newblock \emph{arXiv preprint arXiv:1808.05326}, 2018.

\bibitem[Zeng et~al.(2020)Zeng, Li, Zhai, and Zhang]{zeng2020counterfactual}
Xiangji Zeng, Yunliang Li, Yuchen Zhai, and Yin Zhang.
\newblock Counterfactual generator: A weakly-supervised method for named entity
  recognition.
\newblock In \emph{Proceedings of the 2020 Conference on Empirical Methods in
  Natural Language Processing (EMNLP)}, pages 7270--7280, 2020.

\bibitem[Zhang et~al.(2019)Zhang, Baldridge, and He]{zhang2019paws}
Yuan Zhang, Jason Baldridge, and Luheng He.
\newblock Paws: Paraphrase adversaries from word scrambling.
\newblock \emph{arXiv preprint arXiv:1904.01130}, 2019.

\end{thebibliography}
\bibliographystyle{plainnat}

\newpage

\appendix

{\huge Appendix}

\section{Published and reproduced models}
We reproduce the Stack-Prop+BERT and Bi-RNN models. The resulting trained models obtain similar results to the published, as shown in Appendix Table \ref{tab:published_scores}.

\begin{table}[th]
    \begin{savenotes}
    \renewcommand{\arraystretch}{0.9}
    \setlength{\tabcolsep}{3pt}
        \centering
        \begin{tabular}{l|ll|ll|ll}
        \toprule
        \multirow{2}{*}{Test Set}  & \multicolumn{2}{c}{\bf ATIS} &\multicolumn{2}{c}{\bf SNIPS} &\multicolumn{2}{c}{\bf NLU-ED} \\
        & Slot & Int.& Slot & Int. \\
        \midrule
        \multicolumn{7}{c}{Stack-Prop+BERT} \\
        \midrule
        Published & 96.1 & 97.5 & 97.0 & 99.0 & na & na \\
        Reproduced & 95.7 & 96.5 & 95.0 & 98.2 & 74.0 & 85.1 \\
        \midrule
        \multicolumn{7}{c}{Bi-RNN} \\
        \midrule
        Published & 94.9 & 97.6 & 89.4* & 97.1* & na & na  \\
        Reproduced & 95.7 & 96.5 & 95.0 & 98.3 & 65.8 & 78.8  \\
        \bottomrule
        \end{tabular}
        \caption{Published and reproduced SF and ID results. The numbers with * indicate that the scores were not published in the original Wang et al. Bi-RNN paper but in the Qin et a. Stack-Prop+BERT article.}
        \label{tab:published_scores}
    \end{savenotes}
\end{table}

\section{External perturbation-based techniques}
\label{sec:Augmented Training Sets}

Most DA techniques focus on modifying data to obtain a semantically valid output. The NATURE operators are designed not only to have a semantically valid output but to maintain the same token-level labels as the original data. This small distinction makes a great difference to the end result and we show in Table \ref{tab:da_res} that the DA techniques are not sufficient to cancel out NATURE's alterations.
To this end, we apply standard DA strategies to the train and validation sets, re-train the model from scratch and illustrate their impact on the model's generalization ability. We use common automatic DA strategies from the NLPaug library \footnote{\url{https://github.com/makcedward/nlpaug}} that allow to easily relabel the augmented data using the original labels. We describe these strategies in Appendix Table \ref{tab:da_describ}.

\begin{savenotes}
\begin{table*}[th]
\renewcommand{\arraystretch}{0.9}
\setlength{\tabcolsep}{3pt}
    \small
    \centering
    \begin{tabular}{p{2.5cm}p{5cm}p{5cm}}
    \toprule
  DA strategy name & Description & Example \\
  \midrule
  \textbf{Keyboard Augmentation} & Simulates keyboard distance error. & \textit{find a tv \textbf{seriSs} called \textbf{armaRdvdon} summer} \\
  \midrule
  \textbf{Spelling Augmentation} & Substitutes word according to spelling mistake dictionary. & \textit{\textbf{fine} a tv \textbf{serie} called armageddon summer} \\
  \midrule
  \textbf{Synonym Augmentation} & Substitutes similar word according to WordNet/PPDB synonym. & \textit{find a tv \textbf{set} series called armageddon summertime} \\
  \midrule
  \textbf{Antonym Augmentation} & Substitutes opposite meaning word according to WordNet antonym. & \textit{\textbf{lose} a tv series called armageddon summer} \\
  \midrule
  \textbf{TF-IDF Augmentation} & Uses the TF-IDF measure to find out how a word should be augmented. & \textit{find tv series called armageddon \textbf{forms}} \\
  \midrule
  \textbf{Contextual Word Embeddings Augmentation} & Feeds surroundings word to BERT, DistilBERT, RoBERTa or XLNet language model to find out the most suitable word for augmentation. & \textit{find a \textbf{second} series called armageddon \textbf{ii}} \\
  \bottomrule
\end{tabular}
\end{table*}
\label{tab:da_describ}
\end{savenotes}

We apply the DA strategies exclusively to the train and validation sets, choosing 1 of the 6 DA functions at random and adding one output to the original dataset which results in a training and validation data twice as large as the original training and validation sets.

\begin{table}[!thbp]
    \centering
    \resizebox{\textwidth}{!}{
        \begin{tabular}{l|ll|ll|ll||ll}
            \toprule
            \multirow{2}{*}{\bf Test Set}  & \multicolumn{2}{c|}{\textbf{ATIS}} & \multicolumn{2}{c|}{\textbf{SNIPS}} & \multicolumn{2}{c||}{\textbf{NLU-ED}} &  \multicolumn{2}{c}{\textbf{Avg.}} \\
            & w/o  & w Aug. & w/o  & w Aug. & w/o  & w Aug. & w/o  & w Aug. \\ 
            \midrule
            Orig & 86.2 & 83.3 (-2.9) & 87.9 & 85.3 (-2.6) & 67.8 & 66.2 (-1.6) & 80.6 & 78.3 (-2.3) \\
            Rand & 66.5 & 69.2 (+2.7) & 39.0 & 48.2 (+9.2) & 56.8 & 56.7  (-0.1) & 54.1 & 58.3  (+4.2) \\
            Hard & 34.9 & 54.0 (+19.1) & 12.9 & 27.1  (+15.2) & 38.9 & 40.7 (+1.8) & 28.9 & 40.6 (+11.7) \\
            \bottomrule
    \end{tabular}
}
\caption{End-to-End (E2E) scores of Stack-Prop+BERT models trained on ATIS, SNIPS and NLU-ED original (w/o) and augmented (w) training data. Each model is evaluated on its respective original, Rand, and Hard test set. We report the unweighted average of the 3 datasets.}
\label{tab:da_res}

\end{table}

We have shown that state-of-the-art SF and ID models do suffer when small perturbations are introduced to the test data. We now run experiments on augmented data in order to test the models' performances on larger and slightly more diverse train sets.
Table~\ref{tab:da_res} reports E2E scores of Stack-Prop+BERT~\footnote{Performances of the Bi-RNN model show very similar trends.} model when trained without (w/o) and with (w Aug) data-augmented train and validation sets. Similar to the results table in the main article, we evaluate the model on the Original, Rand, and Hard test sets of ATIS, SNIPS and NLU-ED while also reporting the unweighted average score.\\ 
On one hand, we observe significant gains on the altered test sets (except on NLU-ED Rand) across all benchmarks. The largest increase in performances are obtained on the Hard sets with 19.1\% and 15.2\% of gain on ATIS and SNIPS respectively. The gain can be partially explained by the augmentation of training data size, forcing the model to better generalize and also to the fact that our operator shares some characteristics with the used DA toolkit (i.e., Synonymy).

On the other hand, the performances decrease on the 3 benchmark, by an average of 2.3\%, when the model is evaluated on the Original test sets. DA is a valid strategy in NLP, specially for small sized datasets. However, even the large and more diverse NLU-ED benchmark shows only small improvement and does not solve the unobserved pattern problem exemplified by the NATURE operators. This is a strong indicator that the problem is far from solved, and that there is much room for research.

\section{Qualitative Evaluation}
\label{sec:survey}
In Appendix Tables \ref{fig:survey_instructions} and \ref{fig:survey_excerpt} We show the instructions and an excerpt of the sentences, as presented to the surveyed participants\footnote{We asked the participants to rate the fluency of each utterance (from 1 to 5) in order to average it over the control utterances. Allowing us to establish the annotator capacity of our volunteer participants. We expected this metric to reflect the high quality of the cherry-picked control utterances. As expected, our participants score remained between 4.2 and 5 out of 5.}.

\begin{figure}[th]
\centering
\begin{subfigure}{0.44\textwidth}
  \centering
  \includegraphics[width=1.0\linewidth]{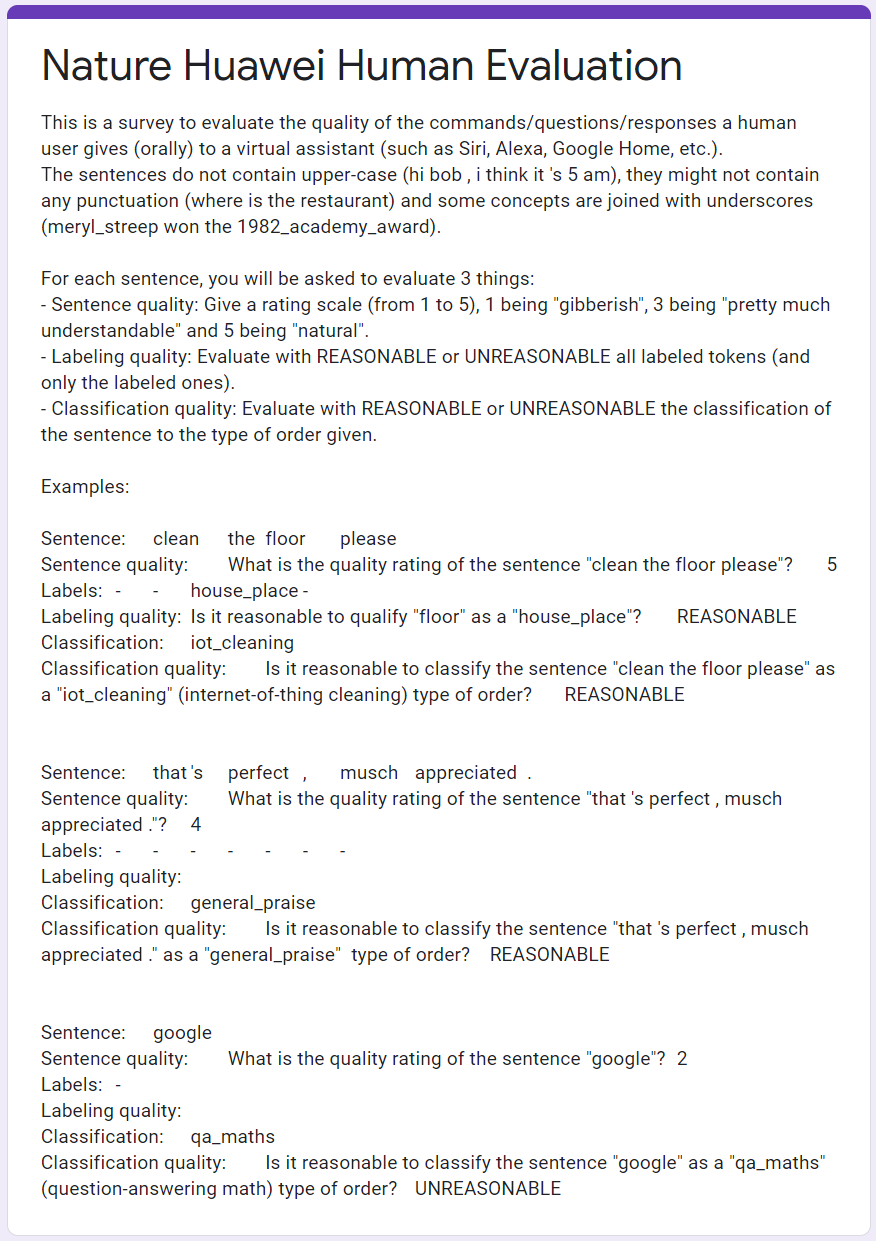}
  \caption{Print-screen of the survey instructions.}
  \label{fig:survey_instructions}
\end{subfigure}%
\begin{subfigure}{0.56\textwidth}
\begin{subfigure}{0.50\textwidth}
  \centering
  \includegraphics[width=0.99\linewidth]{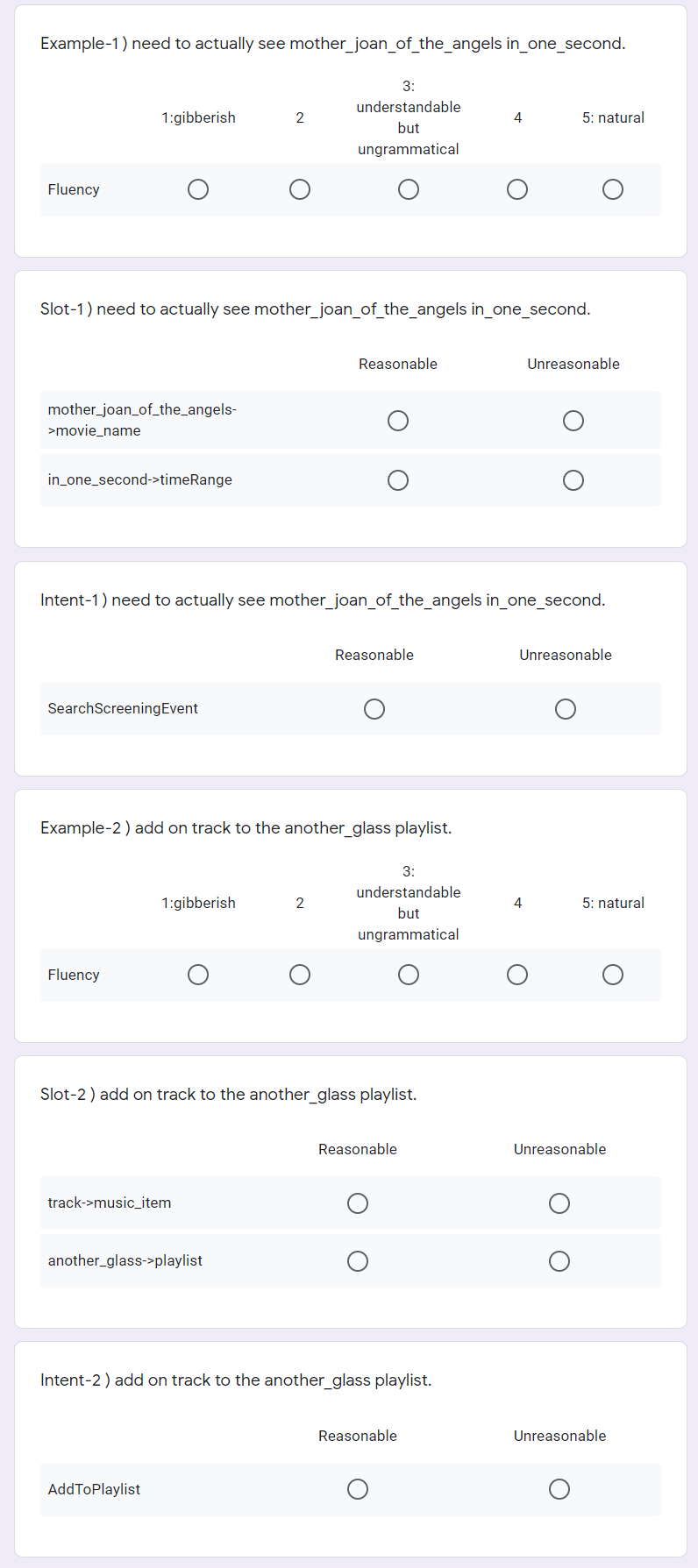}
\end{subfigure}%
\begin{subfigure}{0.50\textwidth}
  \centering
  \includegraphics[width=1.0\linewidth]{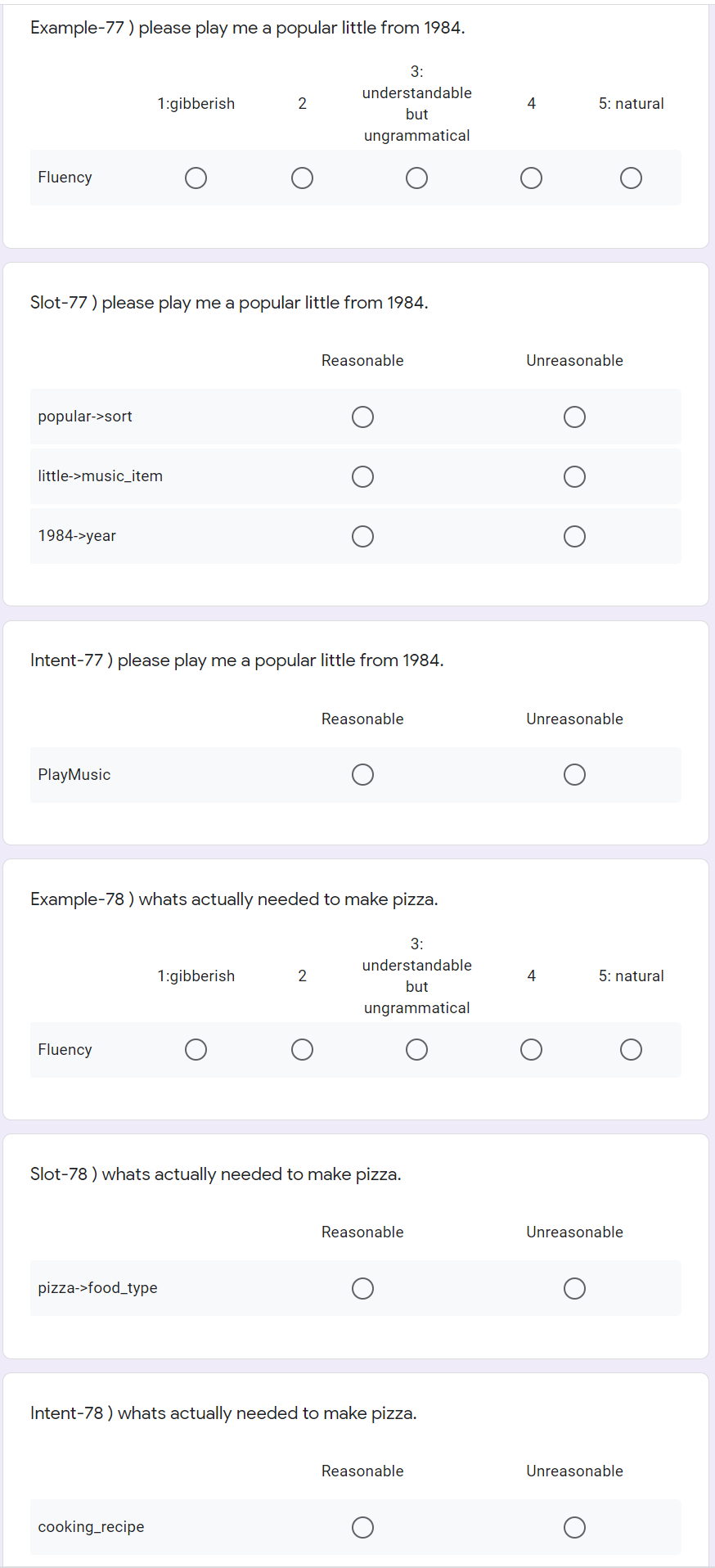}
\end{subfigure}
\caption{Print-screen excerpts of the survey.}
\label{fig:survey_excerpt}
\end{subfigure}
\end{figure}

\begin{table}[th]
    \begin{savenotes}
    \renewcommand{\arraystretch}{0.9}
    \setlength{\tabcolsep}{3pt}
        \centering
        \begin{tabular}{p{2cm}|llllllllllllll}
        \toprule
        Group & \multicolumn{7}{c|}{1} & \multicolumn{7}{c}{2} \\
        Participant Id & 1 & 2 & 3 & 4 & 5 & 6 & 7 & 8 & 9 & 10 & 11 & 12 & 13 & 14 \\
        \midrule
        \multicolumn{15}{c}{Experiment} \\
        \midrule
        Slot & 95.3 & 96.9 & 95.3 & 91.3 & 94.5 & 96.1 & 92.1 & 86.1 & 98.3 & 98.2 & 95.7 & 90.4 & 97.4 & 90.4 \\
        Intent & 83.3 & 93.3 & 87.9 & 83.3 & 90.0 & 91.7 & 93.3 & 76.7 & 90.0 & 88.1 & 93.2 & 87.7 & 84.5 & 81.4 \\
        \midrule
        \multicolumn{15}{c}{Control} \\
        \midrule
        Fluency & 4.9 & 5 & 4.8 & 4.6 & 4.9 & 4.7 & 4.5 & 4.2 & 4.3 & 5 & 5 & 4.9 & 4.8 & 4.2 \\
        Slot & 89.5 & 89.5 & 100 & 94.7 & 100 & 94.7 & 89.5 & 94.7 & 100 & 100 & 100 & 100 & 94.7 & 89.5 \\
        Intent & 91.7 & 100 & 100 & 100 & 100 & 100 & 91.7 & 91.7 & 100 & 100 & 100 & 90.9 & 100 & 100 \\
        \bottomrule
        \end{tabular}
        \caption{Survey results and statistics per participant. The average slot score and the average intent score appear as percentages, the average sentence fluency score appears as a scale from 1 to 5.}
        \label{tab:survey_result_complete}
    \end{savenotes}
\end{table}

\section{Quantitative Evaluation}
In the Appendix Figure\ref{fig:res_summary} we show a more concise illustration of the quantitative experiments' results than Table 7. Appendix Figure\ref{fig:res_summary} shows the E2E score averaged between the benchmarks (ATIS, SNIPS, NLU-ED) and between the two models (Stack-Prop+BERT and Bi-RNN).

\begin{figure}[ht]
    \centering
\begin{minipage}[c]{0.50\textwidth}
    \includegraphics[width=\textwidth]{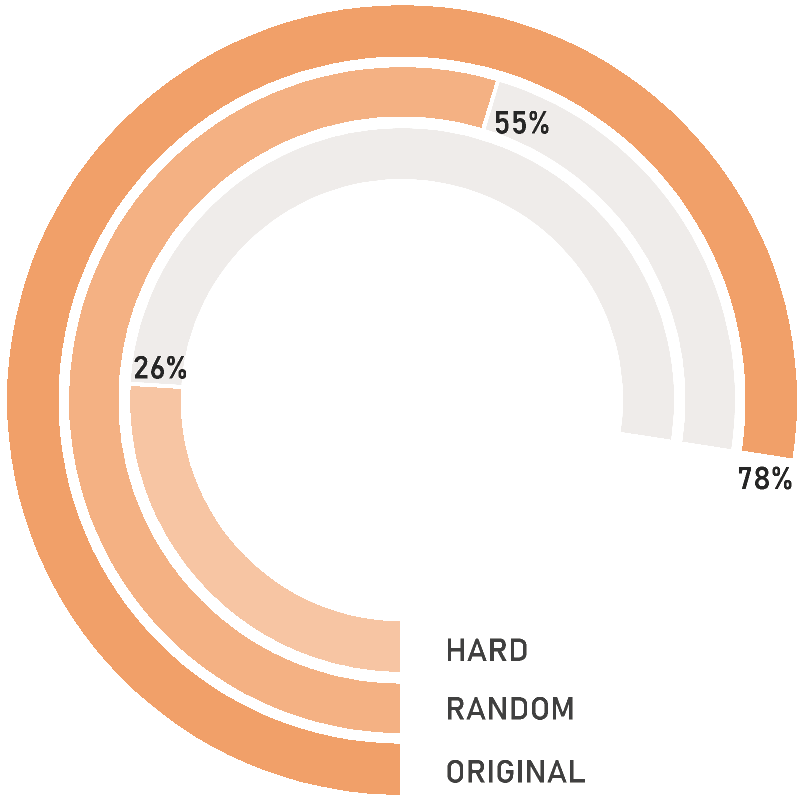}
    \subcaption{Hard and Random scores.}
        \label{fig:hard_rand}
\end{minipage}%
\begin{minipage}[c]{0.50\textwidth}
    \includegraphics[width=\textwidth]{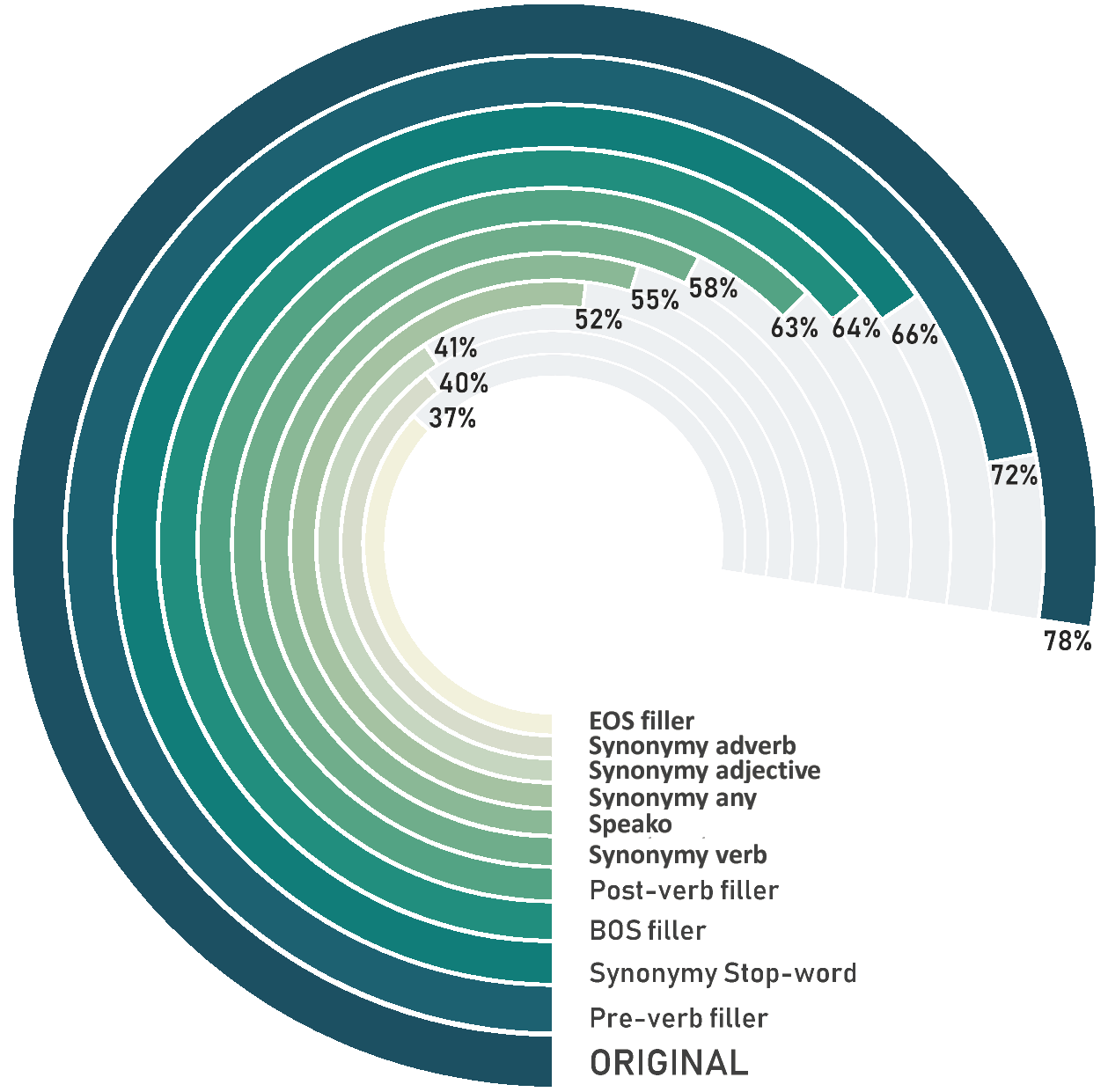}
    \subcaption{Individual operator scores.}
        \label{fig:oper}
\end{minipage}
    \caption{Unweighted average End-to-End score performances averaged between benchmarks (ATIS, SNIPS, and NLU-ED) and models (Stack-Prop+BERT and Bi-RNN). The models were trained using their original train and validation sets and evaluated on altered test sets. Figure~\ref{fig:hard_rand} shows the scores for the Random and Hard evaluation sets while Figure~\ref{fig:oper} shows the scores on 10 evaluation sets, each perturbed with a single NATURE operator, where one operator is applied once to each utterance of the evaluation set}.
    \label{fig:res_summary}
\end{figure}

\section{Complete table of NATURE operators applied to ATIS, SNIPS and NLU-ED}
In the Appendix Table \ref{tab:res_complete} we present all obtained scores ran on 2 models trained on the original train and validation sets of ATIS, SNIPS and NLU-ED and evaluated on the original, random and hard altered test sets.

\begin{savenotes}
\begin{table*}[th]
\renewcommand{\arraystretch}{0.9}
\setlength{\tabcolsep}{3pt}
    \small
    \centering
    \begin{tabular}{l|lll|lll|lll}
    \toprule
  \multirow{2}{*}{Test Set}  & \multicolumn{3}{c}{\textbf{ATIS}} & \multicolumn{3}{c}{\textbf{SNIPS}} & \multicolumn{3}{c}{\textbf{NLU-ED}} \\
  & \vtop{\hbox{\strut Slot}\hbox{\strut (F1)}} & \vtop{\hbox{\strut Intent}\hbox{\strut (Acc)}} & \vtop{\hbox{\strut E2E}\hbox{\strut (Acc)}} & \vtop{\hbox{\strut Slot}\hbox{\strut (F1)}} & \vtop{\hbox{\strut Intent}\hbox{\strut (Acc)}} &  \vtop{\hbox{\strut E2E}\hbox{\strut (Acc)}} & \vtop{\hbox{\strut Slot}\hbox{\strut (F1)}} & \vtop{\hbox{\strut Intent}\hbox{\strut (Acc)}} &  \vtop{\hbox{\strut E2E}\hbox{\strut (Acc)}} \\
  \midrule
  \multicolumn{10}{c}{Stack-Prop+BERT} \\
  \midrule
  Original & 95.7 & 96.5 & 86.2 & 95.0 & 98.3 & 87.9 & 74.0 & 85.1 & 67.8 \\
  Random & 91.3 & 95.0 & 66.5 & 83.4 & 96.1 & 53.8 & 67.4 & 76.1 & 56.8 \\
  & \quad $\pm$ 0.1 & \quad $\pm$ 0.3 & \quad $\pm$ 1.0 & \quad $\pm$ 0.5 & \quad $\pm$ 0.3 & \quad $\pm$ 3.2 & \quad $\pm$ 0.1 & \quad $\pm$ 0.2 & \quad $\pm$ 0.2 \\
  Hard & 82.3 & 90.7 & 34.9 & 70.6 & 95.3 & 12.9 & 55.5 & 62.7 & 38.9 \\
  \midrule
  \multicolumn{10}{c}{Bi-RNN} \\
  \midrule
  Original & 94.7 & 97.6 & 84.3 & 88.9 & 97.6 & 77.3 & 65.9 & 82.1 & 61.9 \\
  Random & 89.9 & 94.3 & 61.8 & 75.6 & 94.1 & 39.0 & 60.6 & 70.8 & 50.1 \\
  & \quad $\pm$ 0.1 & \quad $\pm$ 0.1 & \quad $\pm$ 1.6 & \quad $\pm$ 0.5 & \quad $\pm$ 0.1 & \quad $\pm$ 2.5 & \quad $\pm$ 0.4 & \quad $\pm$ 0.4 & \quad $\pm$ 0.3 \\
  Hard & 79.9 & 92.0 & 27.6 & 62.4 & 92.9 & 7.0 & 49.6 & 58.8 & 34.5 \\
  \bottomrule
\end{tabular}
\caption{Stack-Prop+BERT and Bi-RNN performances for ATIS, SNIPS and NLU-ED. We report F1 slot filling, accuracy for intent detection and end-to-end accuracy overall. The reported scores of the Random altered test set are a mean of 10 random distribution of processes and is accompanied by the variance score.}
\label{tab:res_complete}
\end{table*}
\end{savenotes}

\section{Manual analysis of utterance weight}
\begin{figure}
    \begin{minipage}{.48\textwidth}
        \centering
        \includegraphics[width=42mm,scale=0.5]{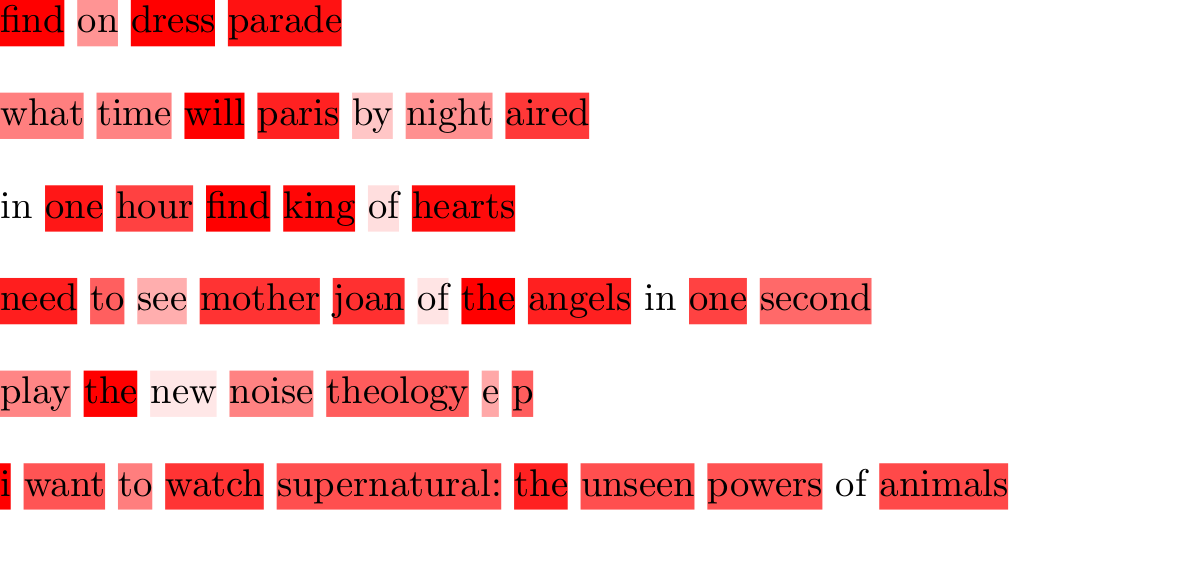}
        \subcaption{Heat-map of original SNIPS utterances.}
        \label{fig:heatmap_snips}
    \end{minipage}
    \begin{minipage}{.48\textwidth}
        \centering
        \includegraphics[width=42mm,scale=0.5]{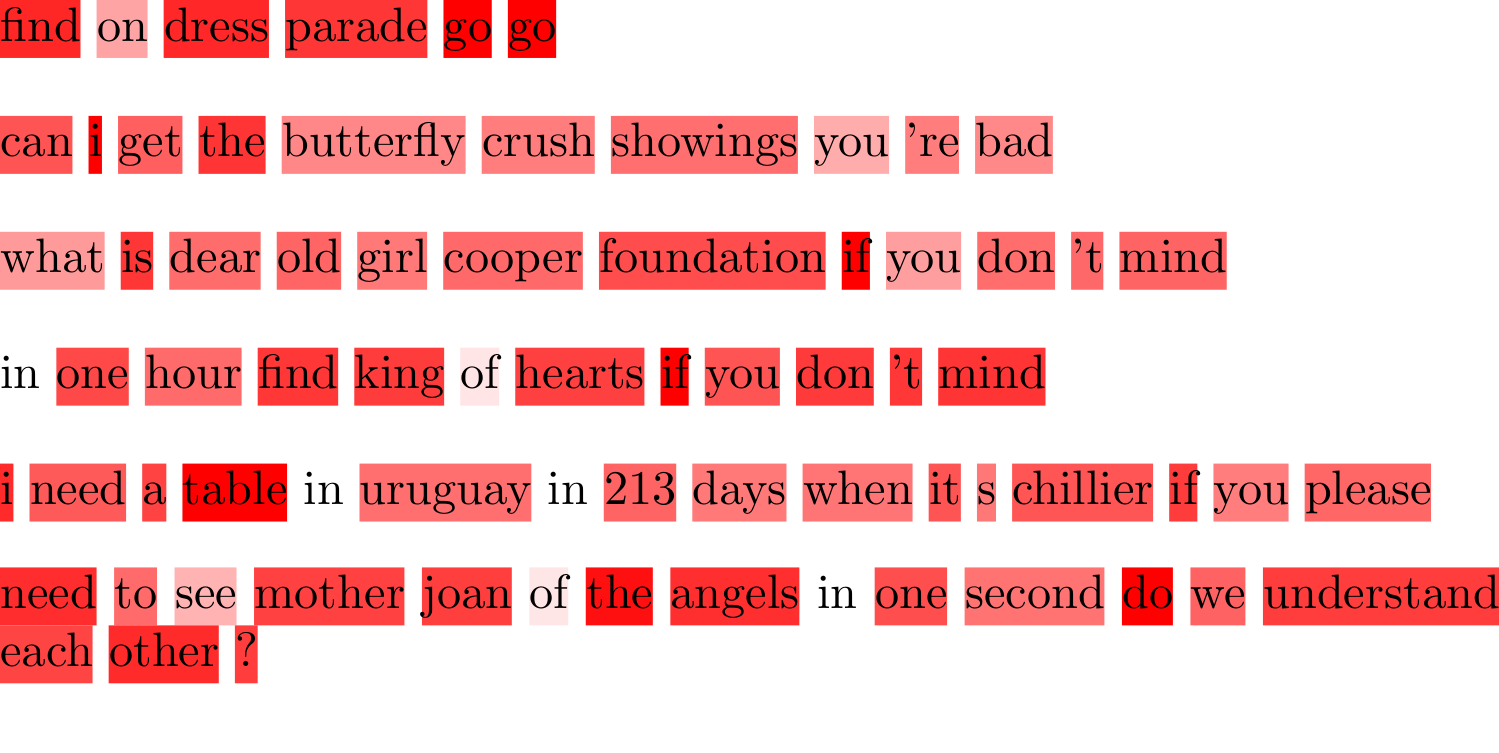}
        \subcaption{Heat-map of EOS filler-altered utterances.}
        \label{fig:heatmap_fill}
    \end{minipage}
    \begin{minipage}{.48\textwidth}
        \centering
        \includegraphics[width=42mm,scale=0.5]{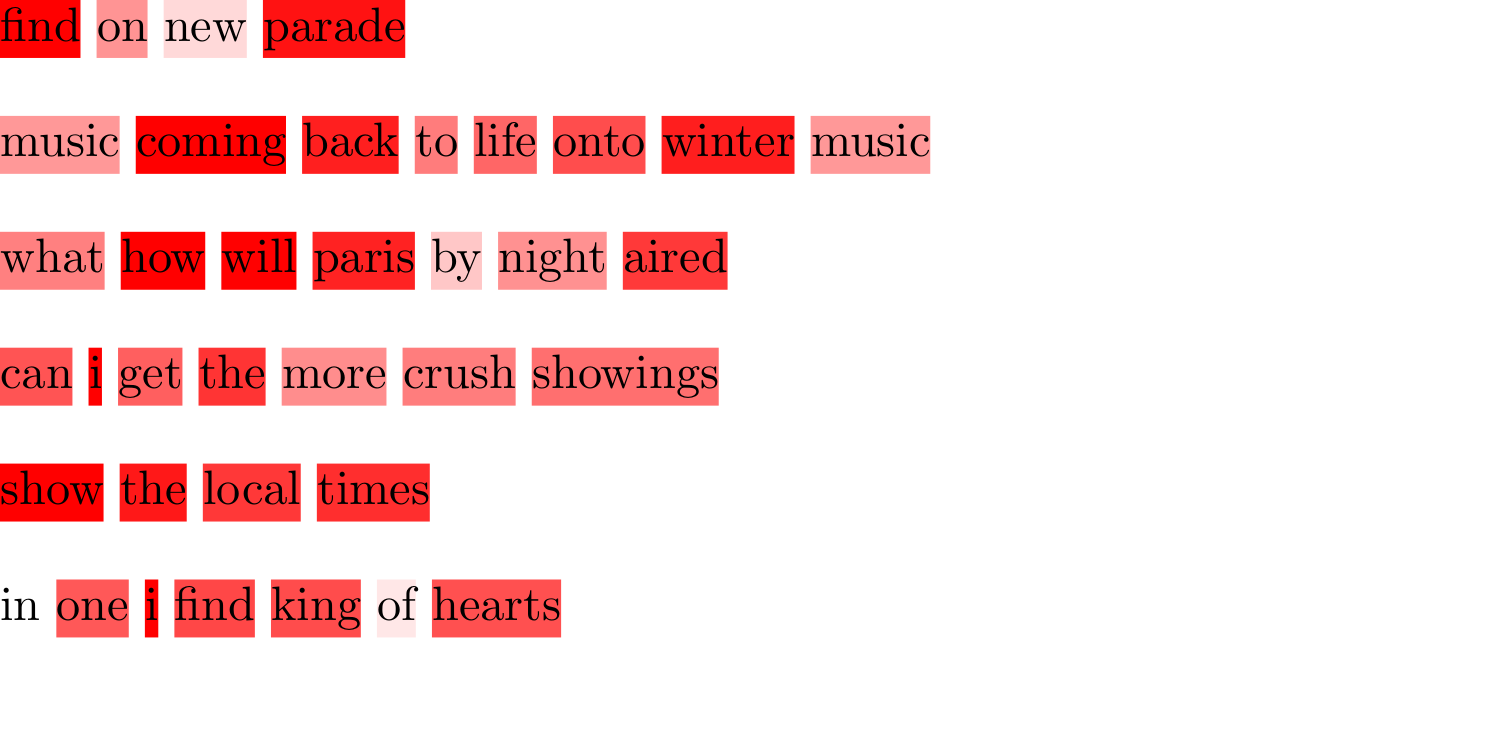}
        \subcaption{Heat-map of Synonymy Adjective-altered utterances.}
        \label{fig:heatmap_syn}
    \end{minipage}
    \begin{minipage}{.48\textwidth}
        \centering
        \includegraphics[width=42mm,scale=0.5]{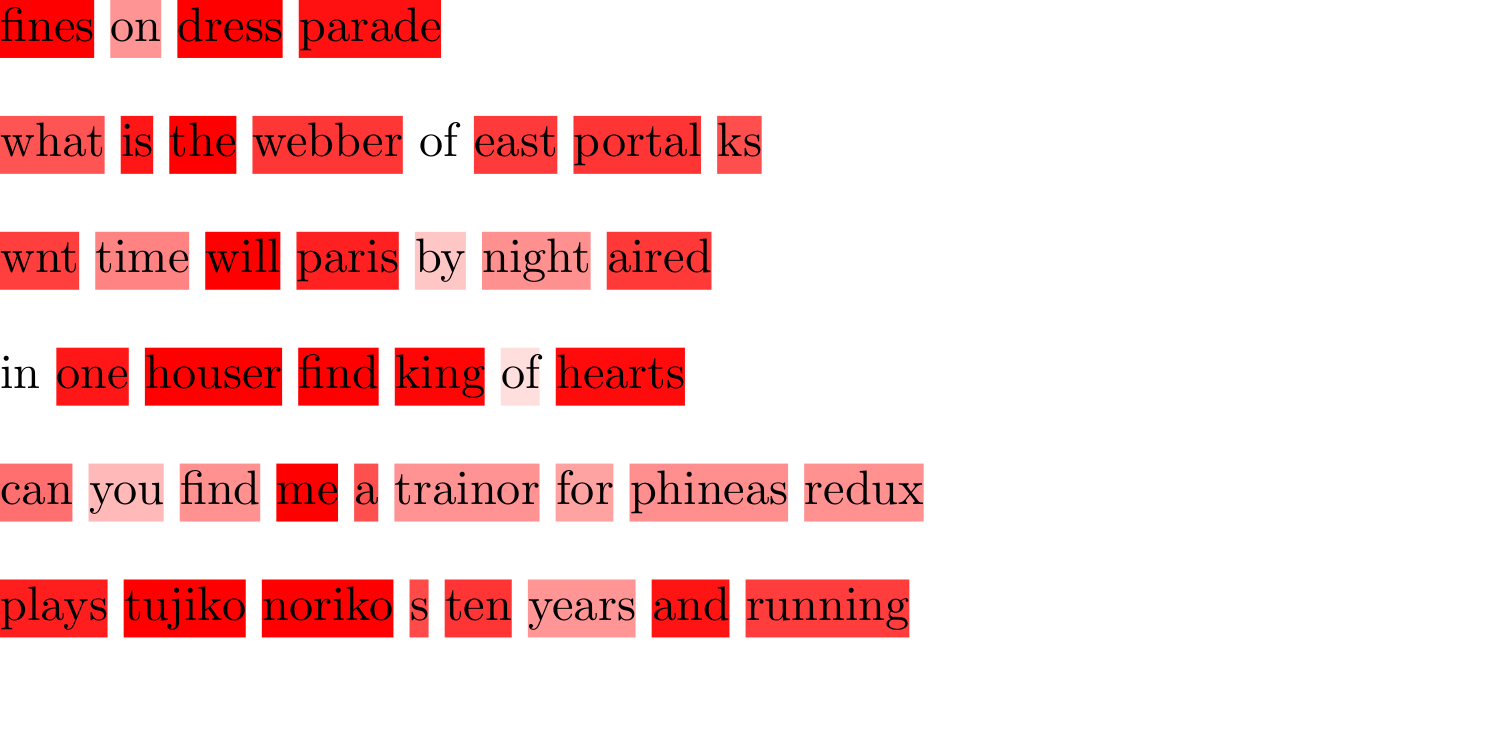}
        \subcaption{Heat-map of Speako-altered utterances.}
        \label{fig:heatmap_speako}
    \end{minipage}
    \caption{Heat-maps of SNIPS utterances whose SF and ID labels were wrongly predicted by the Stack-Prop+BERT model. The more intense the color, the greater the LIME weight.}
    \label{fig:heatmaps}
\end{figure}

To better understand the underlying processes of the state-of-the-art models, we use the LIME tool\footnote{\url{https://github.com/marcotcr/lime}} to produce and analyze multiple self-attention weight heat-maps. This process allows us to better understand what tokens the models focus best to make their prediction. In Figure \ref{fig:heatmaps} we show a representative excerpt heat-maps for wrongly predicted sentences (for both SF and ID). One for the unchanged SNIPS test set and one for each type of operator. At first sight, we notice that the attention is quite evenly distributed among all tokens in the sentence. However, if we carefully examine the small differences between them, we observe a tendency to often focus more heavily on verbs, nouns, certain types of stop words (such as \textit{"the"}) as well as tokens appearing at the extremities of the utterance (although it might not be immediately evident in these small samples of small utterances). It also shows that higher attention is given to verbs and certain stop words at the end of the sentence. This is evident in all Figures but particularly in Figure \ref{fig:heatmap_fill}, where we can see high attention on non-frequent tokens (for the benchmark), such as \textit{"if"} or \textit{"?"}. After more careful analysis, we observe in Figure \ref{fig:heatmap_fill} that the attention is often high for the added filler. This is not the case in Figure \ref{fig:heatmap_syn}, where the attention of the altered synonym is usually low if it doesn't replace a noun. As for the Figure \ref{fig:heatmap_speako}, if we take for example the utterance \textit{what time will paris by night aired}, we observe that just as for the original utterance (and the Synonymy Adjective-altered) the self attention is just as high in the tokens \textit{will}, \textit{paris} and \textit{aired} but it also introduces a high weight on the Speako altered token \textit{want $\xrightarrow{}$ \textit{wnt}}, which doesn't appear in the original utterance.

\section{Complete NATURE operators applied to Data Augmented versions of ATIS, SNIPS and NLU-ED}
In the Appendix Table \ref{tab:da_res_complete} we present all obtained scores ran on 2 models trained on a Data Augmented version of ATIS, SNIPS and NLU-ED.

\begin{savenotes}
  \renewcommand{\arraystretch}{1.4}
  \setlength{\tabcolsep}{3pt}
    \centering
    \begin{table}
        \begin{tabular}{p{0.07\textwidth}|p{0.45\textwidth}|p{0.05\textwidth}|p{0.43\textwidth}}
        \toprule
        \multicolumn{4}{c}{\textbf{Original:} find a tv series called armageddon summer} \\
        \midrule
        \multicolumn{2}{c|}{\textbf{NATURE}} & \multicolumn{2}{c}{\textbf{DA}} \\
        \midrule
        BOS Filler & \textbf{yeah so} find a tv series called armageddon summer & Keyb. & find a tv \textbf{seriSs} called \textbf{armaRdvdon} summer \\
        PreV Filler & \textbf{basically} find a tv series called armageddon summer & Spell. & \textbf{fine} a tv \textbf{serie} called armageddon summer \\
        PosV Filler & find \textbf{you know} a tv series called armageddon summer & Syn. & find a tv \textbf{set} series called armageddon summertime \\
        EOS Filler & find a tv series called armageddon summer \textbf{if it pleases mi liege} & Ant. & \textbf{lose} a tv series called armageddon summer \\
        Syn. V. & \textbf{finds} a tv series called armageddon summer & TF IDF & find tv series called armageddon \textbf{forms} \\
        Syn. Adj. & find a tv series called \textbf{last} summer & Ctxt. WE. & find a \textbf{second} series called armageddon \textbf{ii} \\
        Syn. Adv. & find a \textbf{another} series called armageddon summer & & \\
        Syn. SW & find \textbf{and} tv series called armageddon summer & & \\
        Speako & find a tv \textbf{serie} called armageddon summer & & \\
        \bottomrule
        \end{tabular}
    \caption{Nature and DA candidates for the same utterance.}
    \label{fig:nat_vs_da}
    \end{table}
\end{savenotes}

\begin{savenotes}
\begin{table*}[th]
\renewcommand{\arraystretch}{0.9}
\setlength{\tabcolsep}{3pt}
    \small
    \centering
    \begin{tabular}{l|lll|lll|lll}
    \toprule
  \multirow{2}{*}{Test Set}  & \multicolumn{3}{c}{\textbf{ATIS}} & \multicolumn{3}{c}{\textbf{SNIPS}} & \multicolumn{3}{c}{\textbf{NLU-ED}} \\
  & \vtop{\hbox{\strut Slot}\hbox{\strut (F1)}} & \vtop{\hbox{\strut Intent}\hbox{\strut (Acc)}} & \vtop{\hbox{\strut E2E}\hbox{\strut (Acc)}} & \vtop{\hbox{\strut Slot}\hbox{\strut (F1)}} & \vtop{\hbox{\strut Intent}\hbox{\strut (Acc)}} &  \vtop{\hbox{\strut E2E}\hbox{\strut (Acc)}} & \vtop{\hbox{\strut Slot}\hbox{\strut (F1)}} & \vtop{\hbox{\strut Intent}\hbox{\strut (Acc)}} &  \vtop{\hbox{\strut E2E}\hbox{\strut (Acc)}} \\
  \midrule
  \multicolumn{10}{c}{Stack-Prop+BERT} \\
  \midrule
  Original & 94.7 & 95.7 & 83.3 & 93.8 & 97.7 & 85.3 & 72.4 & 83.8 & 66.2 \\
  Random & 91.7 & 94.3 & 69.2 & 85.7 & 96.0 & 64.4 & 67.3 & 75.6 & 56.7 \\
   & \quad $\pm$ 0.0 & \quad $\pm$ 0.1 & \quad $\pm$ 0.9 & \quad $\pm$ 0.2 & \quad $\pm$ 0.4 & \quad $\pm$ 1.5 & \quad $\pm$ 0.2 & \quad $\pm$ 0.1 & \quad $\pm$ 0.2 \\
  Hard & 87.2 & 91.0 & 54.0 & 72.7 & 95.1 & 27.1 & 55.3 & 64.0 & 40.7 \\
  \midrule
  \multicolumn{10}{c}{Bi-RNN} \\
  \midrule
  Original & 93.7 & 96.9 & 81.8 & 86.2 & 97.6 & 69.7 & 66.3 & 82.5 & 61.8 \\
  Random & 90.3 & 93.9 & 65.6 & 77.4 & 95.3 & 48.2 & 61.2 & 73.4 & 51.8 \\
  Random & \quad $\pm$ 0.1 & \quad $\pm$ 0.2 & \quad $\pm$ 1.1 & \quad $\pm$ 0.3 & \quad $\pm$ 0.2 & \quad $\pm$ 1.8 & \quad $\pm$ 0.1 & \quad $\pm$ 0.2 & \quad $\pm$ 0.2 \\
  Hard & 83.2 & 92.8 & 43.0 & 65.0 & 94.1 & 19.1 & 62.1 & 50.2 & 38.6 \\
  \bottomrule
\end{tabular}
\caption{Stack-Prop+BERT and Bi-RNN performances for ATIS, SNIPS and NLU-ED using data augmentation on the train and validation sets. We report F1 slot filling, accuracy for intent detection and end-to-end accuracy overall. The reported scores of the Random altered test set are a mean of 10 random distribution of processes and is accompanied by the variance score.}
\label{tab:da_res_complete}
\end{table*}
\end{savenotes}

\end{document}